# SpinCastML—An Open Decision-Making Application for Inverse Design of Electrospinning Manufacturing: A Machine Learning, Optimal Sampling, and Inverse Monte Carlo Approach

Elisa Roldán* and Tasneem Sabir

**ABSTRACT:** Electrospinning enables the fabrication of micro- and nanoscale fibrous materials with highly tunable architectures, yet its rational design remains challenging due to the strongly coupled effects of polymer chemistry and concentration, solvent selection, and processing conditions. The process is intrinsically multivariate and nonlinear; small variations in formulation or operating parameters can shift electrohydrodynamic regimes and generate distinct fiber-diameter distributions. Because functional performance often depends on distribution shape and variability rather than on mean diameter alone, electrospinning design must be addressed as a distributional and chemically constrained problem. SpinCastML is a distribution-aware and chemically constrained inverse-design framework implemented as an open standalone executable. The platform is built on a rigorously curated data set of 68,480 individual fiber-diameter measurements extracted from 1,778 data sets spanning 16 polymers. To address high-dimensional heterogeneity and material imbalance, the framework integrates polymer-balanced Sobol and D-optimal sampling with systematic benchmarking of 11 machine-learning algorithms. Rather than predicting a single scalar outcome, SpinCastML models complete fiber-diameter distributions, enabling quantitative assessment of process robustness and probability of meeting user-defined targets. A Cubist model trained under balanced Sobol+D-optimal sampling achieved robust global predictive performance ($R^2 > 0.92$) with stable generalization across chemically diverse systems. The embedded Inverse Monte Carlo engine formulates electrospinning as a probabilistic inverse problem, generating chemically feasible polymer−solvent−process configurations that satisfy compatibility constraints while assigning quantified success probabilities. Experimental distributions are reconstructed with $R^2 > 0.90$, and success rates are predicted within <1% absolute error across independent systems. By integrating distribution-level modeling, chemical feasibility constraints, and probabilistic inverse simulation within a reproducible framework, SpinCastML advances electrospinning from empirical parameter tuning toward quantitatively guided, application-driven design.

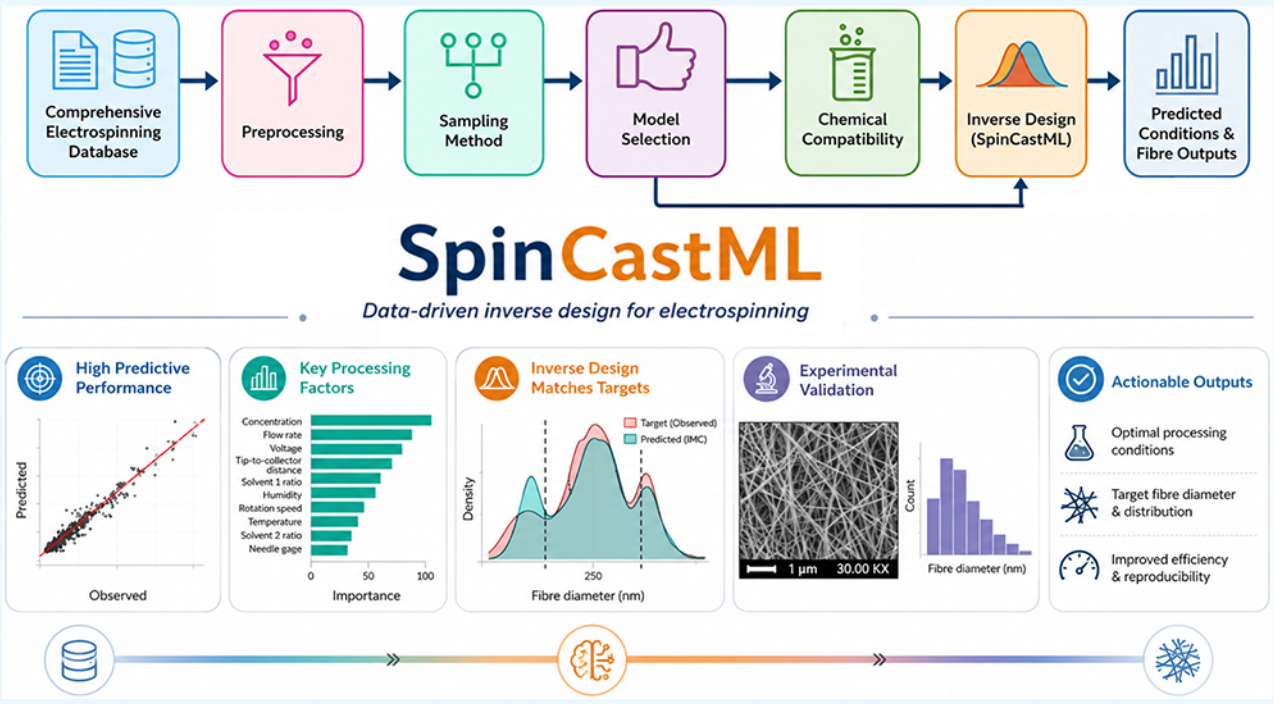

## ■ INTRODUCTION

Electrospinning is a very versatile manufacturing technique capable of producing continuous micro- to nanoscale fibers from polymer solutions or melts by applying a high voltage between a spinneret and a grounded collector.[1] Under an electric field, the polymeric solution elongates into a Taylor cone, launches a charged jet, and undergoes a characteristic whipping (bending) instability that dramatically thins the filament before solidification, yielding nonwoven fibrous mats with very high surface-area-to-volume ratios ideal for multiple purposes.[2] Morphology (fiber diameter, fiber alignment, interfiber separation, porosity, bead formation), topography (surface smoothness/roughness) or mechanical properties of the scaffolds emerge from coupled effects of solution properties (concentration, viscosity, molecular weight, conductivity, surface tension), processing parameters (e.g., voltage, flow rate, needle-collector distance, revolutions of the mandrel), and ambient conditions (temperature, humidity).[3]

Because of its micro- and nanostructure and high surface-area-to-volume ratios, electrospun materials have been adopted across a wide range of applications from healthcare with tissue engineering scaffolds,[4,5] drug and gene delivery,[6−8] wound dressings,[9] hemostatic,[10] or soft bioelectronics and biosensors,[11,12] to air−water filtration and protective membranes,[13,14] transpirable and waterproof textiles,[15] energy and catalysis (e.g., energy storage, energy generator, photocatalysts),[16−18] and food and environmental interfaces.[19] Additionally, case studies of these nanostructures have been highlighted as key metamaterials and metasurfaces for health and well-being, therapeutics and sustainability in a policy report published by the Institution of Mechanical Engineers of the United Kingdom.[20] Application-specific requirements often hinge on achieving target fiber diameter distributions and architectures (e.g., aligned vs random fibers).[21] This is a key aspect for biomimicry where native tissues frequently rely on characteristic bimodal collagen-fiber patterns often lost upon injury,[21−24] underscores how nanofiber organization directly shapes mechanical behavior, permeability, surface chemistry, and biological responses.

Despite decades of use of electrospinning technique, predicting *a priori* what morphology will be obtained for a given formulation and setting remains difficult. To our knowledge, no prior studies have reported the settings for the electrospinner given a desired morphology (inverse design). The difficulty arises because the process is strongly multivariate and nonlinear; and small changes in solution or ambient conditions can shift regimes (e.g., from obtaining microfibers to nanofibers, or from beaded fibers to nonbeaded fibers). In practice, laboratories still rely heavily on trial-and-error optimization, which is time- and material-intensive, limiting the sustainability of the process.[5] This motivates data-driven approaches that learn mappings from operating conditions to fiber outcomes and can support design-space exploration under realistic constraints.

Recent efforts have applied machine learning (ML) to model electrospinning responses, most commonly fiber diameter, using algorithms such as artificial neural networks,[25] kernel and ensemble methods,[26,27] and hybrid response-surface strategies.[28] Several studies report accurate prediction of diameter and offer sensitivity analyses;[29−32] more recently, two user-facing web platforms have begun to appear to democratize these models to accelerate decision-making,[21,33] being FiberCastML the first open web application able to forward predict not just individual fiber diameter but the full predictive distributions of fibers providing a robust data set and validation.[21] Together, these advances point to a transition from ad-hoc screening toward reproducible, model-assisted design.

Sustainability concerns further amplify the need for informed design. Conventional solvents widely used in electrospinning (e.g., Dichloromethane, Dimethylformamide, 2,2,2-Trifluoroethanol) carry human-health and environmental burdens, and greener solvent systems and recycled/biobased polymers are actively being pursued.[34] Additionally, methods that reduce the number of wet-lab iterations, and that can encode chemical feasibility or "green" constraints during design, can support safer, lower-waste development of electrospun products.

This work advances electrospinning design by introducing SpinCastML, an interactive and reproducible framework, distributed as a standalone executable, for inverse determination of electrospinning formulations and operating conditions that yield targeted nanofiber diameter distributions. Unlike conventional approaches that predict a single mean outcome, SpinCastML is explicitly distribution-aware, capturing the full statistical structure of electrospun fibers and enabling design toward complex, polymer-specific diameter architectures that reflect real experimental variability.

The framework integrates four key methodological elements. First, it leverages a rigorously curated, literature-derived data set comprising 68,480 fiber-diameter observations with transparent preprocessing and provenance tracking. Second, it systematically evaluates structured sampling strategies (including random, Sobol, D-optimal, and polymer-balanced hybrid designs) to address data imbalance, improve coverage of high-dimensional process space, and reduce computational cost. Third, predictive models are selected through cross-validated benchmarking across 11 machine-learning algorithms, yielding a robust global model that generalizes across polymers. Fourth, these components are unified within a chemically constrained Inverse Monte Carlo (IMC) engine that inverts a target fiber diameter and tolerance into families of feasible electrospinning conditions while enforcing polymer−solvent solubility, solvent−solvent miscibility, and user-selectable strictness.

Beyond methodological development, SpinCastML operationalizes inverse electrospinning design in an open and accessible form. The application enables researchers to upload and analyze their own curated data sets without prior machine-learning expertise, providing standardized data schemas, automated preprocessing, transparent model evaluation, and reproducible reporting. This design lowers the technical barrier to advanced modeling while supporting laboratory-specific exploration of process−structure relationships.

Together, SpinCastML establishes a distribution-aware and chemically informed paradigm for inverse electrospinning design, replacing trial-and-error optimization with probabilistic, decision-ready guidance that is reproducible, auditable, and resilient to experimental uncertainty. By reframing electrospinning as a distributional inverse problem and embedding chemical feasibility directly into the design loop, this work sets a new standard for how electrospun processes are designed, validated, and interrogated under realistic manufacturing variability.

## RESULTS AND DISCUSSION

To develop a reproducible framework for inverse electrospinning design, a comprehensive data set was curated. The data set used in this study comprises 68,480 electrospinning records collected from the literature, spanning polymers, solvent systems, operating conditions, and fiber diameters. As is typical in multisource experimental data sets, observations are imbalanced, with polymers overrepresented against others. To prevent dominant polymers from biasing model learning, we compared three sampling strategies (simple random sampling, Sobol+D-optimal sampling, and polymer-balanced Sobol+D-optimal sampling) which together improve domain coverage, reduce redundancy, ensure minority-polymer representation and reduce computational resources. After sampling, data were split into a user-selected rate (default: 70% training and 30% testing), and cross-validated ($k = 3, 5, 10$) across 11 machine-learning algorithms using root mean squared error (RMSE), mean absolute error (MAE), mean absolute percentage error (MAPE), and the coefficient of determination ($R^2$). A single global model was selected because it leverages shared physical relationships across polymers, learns generalizable patterns from all 68,480 records, and substantially reduces variance by borrowing statistical strength from well-sampled polymers. This global model approach is also highly beneficial to understand the production of copolymeric electrospun scaffolds. This yields a coherent and stable representation of electrospinning behavior across materials. The final global model was then integrated into an Inverse Monte Carlo (IMC) framework using quasi-random Sobol/LHS sampling to generate chemically feasible configurations and electrospinning setup and predict full fiber-diameter distributions. Figure 1 outlines the followed process; full description of the methodology can be found in Methods Section.

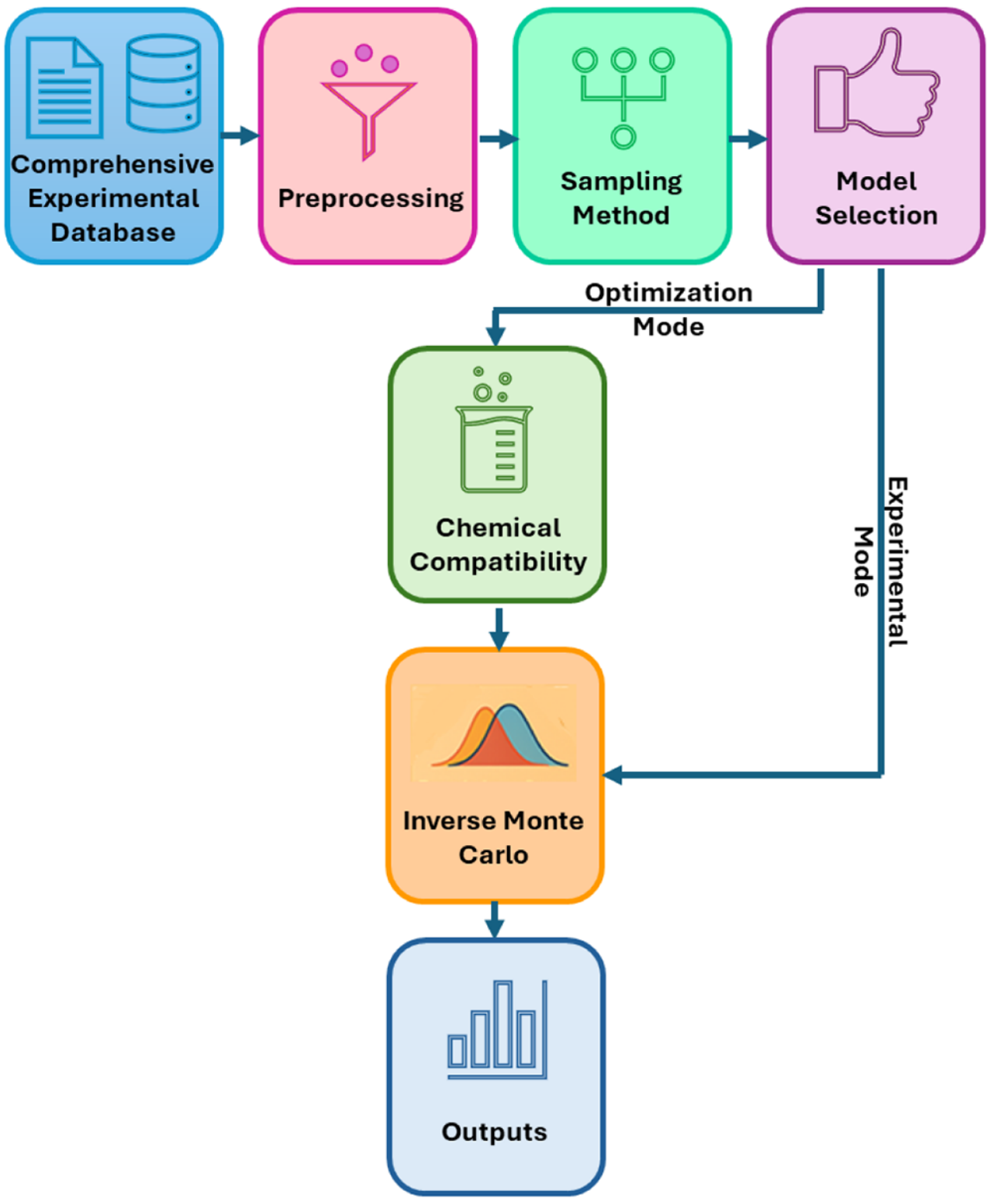

**Figure 1.** Outline of the followed methodology.

## Descriptive Analysis of Electrospun Fiber Diameters

Across all polymers, fiber diameters show a broad, right-skewed and heavy-tailed distribution (median 340 nm, IQR 199−664 nm, mean 603 nm, skewness 5.82, kurtosis 46.98), indicating that a minority of very thick fibers inflates the mean and variance (Table 1). Stratification by polymer reveals clear, material-specific profiles: Polyether−ether−ketone -sulfonated and polyacrylonitrile yield consistently thin, tightly clustered fibers (medians between 184−202 nm and narrow IQRs), whereas polyvinylidene fluoride and polyvinylpyrrolidone exhibit pronounced upper tails (skewness 4.51 and 4.09; kurtosis 24.81 and 19.26, respectively), and polystyrene/poly(D,L-lactide)/poly(methyl methacrylate) occupy the thicker regime with broader spreads. Midrange targets (300−500 nm) are well represented by polylactic acid, polycaprolactone, polyethylene terephthalate, and cellulose acetate. Given the skew, medians and IQRs are more representative than means, and polymer-level summaries prevent cross-material mixing from masking meaningful structure.

The large sample size spanning 1,778 studies, 68,480 fiber-diameter and 16 polymers is a strength: it captures realistic variability inherent to electrospinning (solution properties and process parameters), increases statistical power for between-polymer contrasts, stabilizes robust estimators in heavy-tailed settings, and provides a high-coverage basis for downstream modeling, thereby enhancing generalizability beyond narrowly defined experiments.[35] However, such scale also introduces imbalance between polymers. Using the full data set without correction would overrepresent dominant polymers, under-represent minority ones, distort learned global relationships, and impose unnecessary computational costs. For this reason, SpinCastML applies targeted sampling strategies (random sampling, Sobol+D-optimal sampling, and polymer-balanced Sobol+D-optimal sampling) to control imbalance, reduce redundancy, and preserve both global variability and polymer-specific information in a computationally efficient manner.

## Sampling Method

Three sampling strategies were evaluated as parallel alternatives rather than as a single sequential pipeline. Random sampling was used as a baseline because it preserves the empirical distribution of the original data set but does not guarantee coverage of high-dimensional experimental space or minority-polymer representation. Sobol+D-optimal sampling was evaluated as a structured strategy in which Sobol sequences first generate a space-filling candidate set across the continuous predictor domain, and D-optimal design subsequently selects the most statistically informative subset. Polymer-balanced Sobol+D-optimal sampling extends this approach by applying the structured sampling procedure within polymer strata, thereby combining continuous-domain coverage with improved representation of under-sampled polymers. This design allowed the contribution of each sampling strategy to be evaluated quantitatively in terms of polymer coverage, redundancy reduction, computational efficiency, and model generalization.

**Table 1. Descriptive Analysis of Electrospun Fiber Diameters**

| Polymer | Diameter of the fibers (nm) Mean | Std dev | Q1 (25%) | Median (50%) | Q3 (75%) | Kurtosis | Skewness |
|---|---|---|---|---|---|---|---|
| Cellulose acetate (CA) | 538.889 | 365.794 | 152.007 | 488.195 | 786.400 | −0.790 | 0.501 |
| Gelatin | 368.373 | 308.915 | 131.289 | 258.637 | 528.665 | 3.285 | 1.669 |
| Polyamide-6 (Nylon-6) | 372.606 | 260.577 | 173.296 | 208.958 | 683.719 | −1.630 | 0.473 |
| Polyacrylonitrile (PAN) | 210.454 | 59.539 | 175.479 | 201.735 | 244.625 | 1.020 | 0.794 |
| Polycaprolactone (PCL) | 356.687 | 154.398 | 277.604 | 324.811 | 398.240 | 0.463 | 0.980 |
| Poly(D,L-lactide) (PDLLA) | 839.983 | 788.921 | 198.476 | 633.173 | 1139.350 | −0.070 | 1.033 |
| Polyether−ether−ketone (PEEK-sulfonated) | 187.800 | 21.899 | 170.995 | 183.600 | 200.206 | −0.055 | 0.716 |
| Polyethylene terephthalate (PET) | 404.046 | 173.434 | 212.782 | 399.033 | 530.132 | −1.057 | 0.297 |
| Polylactic acid (PLA) | 378.199 | 161.296 | 247.969 | 308.374 | 498.824 | −0.829 | 0.748 |
| Poly(methyl methacrylate) (PMMA) | 614.207 | 506.856 | 264.908 | 430.438 | 655.330 | 2.058 | 1.603 |
| Polystyrene (PS) | 987.716 | 942.455 | 348.229 | 549.717 | 1386.753 | 2.934 | 1.784 |
| Polyurethane (PU) | 372.917 | 102.391 | 336.332 | 381.183 | 440.890 | −0.800 | −0.266 |
| Polyvinyl alcohol (PVA) | 264.838 | 134.399 | 197.444 | 245.967 | 289.897 | 4.710 | 1.855 |
| Polyvinylidene fluoride (PVDF) | 689.367 | 979.741 | 221.872 | 444.464 | 734.112 | 24.812 | 4.511 |
| Polyvinylpyrrolidone (PVP) | 1102.890 | 1815.495 | 266.956 | 554.793 | 1116.621 | 19.256 | 4.086 |
| Poly-$\gamma$-glutamic acid (Y_PGA) | 315.408 | 97.599 | 248.286 | 293.645 | 346.380 | 1.915 | 1.377 |
| **TOTAL** | **602.963** | **915.674** | **199.469** | **339.846** | **664.291** | **46.979** | **5.823** |

As a baseline approach, simple random sampling was applied. As sample size increases from 2,000 to 10,000 records, test performance improves markedly: MAPE drops from 25.895 to 8.641 nm, RMSE and MAE decrease, and $R^2$ increases from 0.76 to 0.94. At very small N (≤4,000), the large gaps between cross-validation and test errors indicate unstable, optimistic estimates and overfitting, which is consistent with recent work showing that small ML data sets tend to overestimate predictive power and yield poorly reproducible models.[35] Beyond 10,000 samples, performance gains plateau while computational time grows sharply (from ∼2 min at N = 10,000 to >11 min at N = 60,000 with an Intel i5 preprocessor), suggesting a regime of diminishing returns where additional data mainly increase cost rather than accuracy. Similar saturation effects between data size, accuracy, and runtime have been reported in general ML benchmarks and in materials informatics.[36] In our case, N = 10,000 offers a good compromise, delivering near-optimal RMSE/$R^2$ with substantially lower runtime than the full 68,480 record subset (supporting data can be found in Supporting Information). These observations align with recent ML studies similarly highlight that richer data sets improve nanofiber diameter prediction but must be balanced against computational efficiency and experimental practicality.[31,33,37] Although random sampling is computationally efficient and straightforward to implement, no guarantees of coverage across high-dimensional feature spaces are provided, and minority polymer–solvent combinations may be under-sampled. Consequently, all subsequent advanced sampling strategies were benchmarked against this baseline.

To improve coverage of continuous experimental variables (e.g., solution concentration, voltage, flow rate, tip-to-collector distance, environmental conditions), Sobol low-discrepancy sequences and D-optimal experimental design were employed as described in the Methods Section. By using Sobol sampling, quasi-random, deterministically uniform points are produced across the experimental domain.[38] In contrast to purely random draws, Sobol sequences minimize clustering and reduce gaps in high-dimensional input regions, thereby improving model generalization and reducing the risk of extrapolation.[39] Because Sobol sampling alone does not account for collinearity or the intrinsic structure of the data set, a D-optimal experimental design step was integrated. Under a D-optimality criterion, the subset of candidate points that maximizes the determinant of the information matrix is selected, thereby minimizing the generalized variance of model parameter estimates.[40] In this manner, the most informative subset of samples is identified given the multivariate relationships among predictors. We observed that if we started with a 10,000 candidate points, this sampling favors points that are highly orthogonal and maximize explanatory variance across numerical predictors. As a result, the sampled set is dominated by "rich" regions of the experimental space, those spanning eight polymers with extensive variation and wide dispersion. When a model is trained on this restricted yet highly informative subset, the fit becomes exceptionally tight ($R^2$ ≈ 0.997), because the design captures only the most structured, high-signal portion of the domain. However, this comes at a cost: the resulting model becomes overspecialized, losing representativeness for polymers and formulation regimes not covered by the optimal design. In practical terms, it behaves like an instrument calibrated only at the most informative points and then evaluated exclusively within that narrow window. It is worth noting that to be able to obtain 10,000 chemically valid and polymer-diverse samples, an initial request of ∼26,200 candidate records is required, because many D-optimal points are later excluded to preserve full chemical diversity and avoid collapsing the model onto a limited subset of the formulation space.

The hybrid sampling strategy (combining Sobol low-discrepancy sequences, D-optimal experimental design, and polymer-balanced stratification) substantially improved data set representativeness and downstream model performance. Compared with naïve random sampling, Sobol+D-optimal sampling produced a markedly more space-filling distribution of continuous variables, reducing clustering and eliminating gaps across the high-dimensional formulation domain.

Polymer-balanced Sobol+D-optimal sampling successfully corrected extreme frequency imbalances present in the raw data set. Minority polymers increased from <1% representation to predefined target ranges, while dominant polymers were down weighted without loss of internal variability. A summary table of pre- (total n per polymer) and postsampling polymer frequencies for 10000 sample size is shown in Table 2. To visually complement the frequency counts reported in Table 2, Supplementary Figure S1 compares the original polymer distribution with the data sets generated by random sampling, Sobol+D-optimal sampling, and polymer-balanced Sobol+D-optimal sampling, highlighting the improved minority-polymer representation achieved by the balanced strategy.

**Table 2. Pre- and Post-Sampling Polymer Frequencies**

| | Pre-sampling | Post-sampling | | |
|---|---|---|---|---|
| Polymer | Total | Random | Sobol+D-optimal | Balanced |
| CA | 1880 | 277 | 275 | 627 |
| GELATIN | 1680 | 272 | 253 | 620 |
| Nylon-6 | 1600 | 243 | 236 | 618 |
| PAN | 4320 | 632 | 641 | 675 |
| PCL | 1720 | 248 | 251 | 622 |
| PDLLA | 640 | 104 | 90 | 563 |
| PEEK-sulfonated | 1160 | 156 | 167 | 599 |
| PET | 480 | 81 | 67 | 480 |
| PLA | 320 | 55 | 45 | 320 |
| PMMA | 4760 | 712 | 696 | 680 |
| PS | 1800 | 247 | 254 | 624 |
| PU | 680 | 88 | 91 | 567 |
| PVA | 6920 | 1011 | 1037 | 701 |
| PVDF | 34920 | 5058 | 5079 | 1053 |
| PVP | 4880 | 710 | 720 | 681 |
| Y_PGA | 720 | 106 | 98 | 570 |
| **TOTAL** | **68480** | **10000** | **10000** | **10000** |

Through this strategy, the sampled data set was designed to preserve meaningful chemical diversity and to support robust model performance across polymers with different solubilities, solvent compatibilities, and electrospinning behaviors. Categorical variables (including polymer identity and up to three solvents) were sampled directly from the empirical distribution observed in the data. To preserve chemical realism, an optional constraint was applied in which polymer–solvent tuples were required to match combinations previously observed exper-imentally. In this way, generation of physically implausible systems were prevented, while novel regions of the continuous parameter space could still be explored. The combined sampling strategy was designed to meet the methodological demands of high-quality predictive modeling in electrospinning research. Compared with purely random sampling, this approach: ensures broad, uniform exploration of multivariate numerical predictors; prioritizes the most statistically informative samples; maintains chemical and formulation diversity; prevents domination by over-represented polymers; preserves experimentally validated polymer–solvent compatibility constraints; and reduces the risk of model bias and overfitting.

Table 3. Machine Learning Models Metrics Per Sampling Method

| Sampling | Model | RMSE | | | MAE | | | MAPE | | | $R^2$ | | |
|---|---|---|---|---|---|---|---|---|---|---|---|---|---|
| | | Test(30%) | CV(train) | ΔRMSE | Test(30%) | CV(train) | ΔMAE | Test(30%) | CV(train) | ΔMAPE | Test(30%) | CV(train) | $ΔR^2$ |
| Balanced | *Mean Balanced* | *491.19* | *496.36* | *-5.17* | *191.28* | *194.88* | *-3.61* | *34.42* | *33.15* | *1.27* | *0.68* | *0.69* | *-0.0128* |
| | cubist | 260.08 | 263.45 | -3.37 | 50.01 | 52.81 | -2.80 | 7.41 | 7.36 | 0.05 | 0.92 | 0.92 | -0.0020 |
| | earth | 587.00 | 599.15 | -12.15 | 285.47 | 296.51 | -11.04 | 71.97 | 71.60 | 0.38 | 0.59 | 0.59 | -0.0035 |
| | gbm | 296.23 | 291.50 | 4.73 | 115.33 | 121.74 | -6.42 | 12.23 | 12.04 | 0.18 | 0.90 | 0.90 | -0.0069 |
| | glmnet | 697.09 | 709.49 | -12.41 | 309.36 | 308.80 | 0.56 | 33.11 | 32.64 | 0.47 | 0.42 | 0.43 | -0.0086 |
| | knn | 260.81 | 259.03 | 1.78 | 49.59 | 54.46 | -4.87 | 35.22 | 33.71 | 1.51 | 0.92 | 0.92 | -0.0050 |
| | lm | 697.10 | 709.79 | -12.70 | 309.99 | 309.67 | 0.32 | 44.58 | 44.48 | 0.10 | 0.42 | 0.43 | -0.0080 |
| | ranger | 510.86 | 507.72 | 3.14 | 182.59 | 182.98 | -0.39 | 16.62 | 16.55 | 0.06 | 0.72 | 0.75 | -0.0301 |
| | rf | 501.80 | 519.26 | -17.46 | 182.63 | 186.17 | -3.54 | 87.11 | 81.43 | 5.68 | 0.72 | 0.73 | -0.0140 |
| | rpart | 709.94 | 674.01 | 35.93 | 379.83 | 372.30 | 7.53 | 44.50 | 43.48 | 1.03 | 0.40 | 0.48 | -0.0874 |
| | svmRadial | 619.31 | 660.39 | -41.08 | 169.89 | 183.77 | -13.88 | 17.46 | 13.27 | 4.18 | 0.58 | 0.55 | 0.0269 |
| | xgbTree | 262.83 | 266.14 | -3.31 | 69.35 | 74.52 | -5.16 | 8.40 | 8.12 | 0.28 | 0.92 | 0.92 | -0.0019 |
| D-Optimal+Sobol | *D-Optimal+Sobol* | *561.09* | *503.73* | *57.36* | *202.86* | *196.98* | *5.88* | *50.33* | *50.62* | *-0.29* | *0.64* | *0.67* | *-0.0236* |
| | cubist | 289.89 | 250.50 | 39.40 | 55.82 | 53.56 | 2.26 | 7.99 | 9.09 | -1.10 | 0.91 | 0.93 | -0.0124 |
| | earth | 657.34 | 590.02 | 67.33 | 309.33 | 292.71 | 16.62 | 87.04 | 85.32 | 1.72 | 0.55 | 0.58 | -0.0384 |
| | gbm | 315.96 | 265.55 | 50.41 | 118.85 | 113.51 | 5.34 | 31.09 | 29.87 | 1.22 | 0.90 | 0.92 | -0.0197 |
| | glmnet | 757.71 | 703.96 | 53.75 | 317.49 | 302.26 | 15.24 | 77.17 | 76.79 | 0.39 | 0.40 | 0.41 | -0.0123 |
| | knn | 470.11 | 314.74 | 155.37 | 88.56 | 84.75 | 3.82 | 12.52 | 16.05 | -3.53 | 0.77 | 0.88 | -0.1133 |
| | lm | 757.72 | 703.19 | 54.53 | 318.19 | 302.73 | 15.46 | 77.61 | 77.01 | 0.60 | 0.40 | 0.41 | -0.0136 |
| | ranger | 579.37 | 526.13 | 53.24 | 187.20 | 185.69 | 1.51 | 42.42 | 44.70 | -2.28 | 0.68 | 0.71 | -0.0327 |
| | rf | 581.12 | 529.50 | 51.62 | 187.01 | 186.45 | 0.56 | 42.21 | 44.86 | -2.65 | 0.68 | 0.71 | -0.0310 |
| | rpart | 725.62 | 706.59 | 19.03 | 372.85 | 371.79 | 1.06 | 124.22 | 122.06 | 2.15 | 0.45 | 0.41 | 0.0410 |
| | svmRadial | 741.64 | 701.85 | 39.79 | 200.60 | 201.70 | -1.09 | 35.21 | 35.05 | 0.16 | 0.45 | 0.46 | -0.0116 |
| | xgbTree | 295.50 | 249.01 | 46.49 | 75.56 | 71.65 | 3.91 | 16.14 | 16.01 | 0.13 | 0.91 | 0.93 | -0.0150 |
| Random | *Mean Random* | *538.64* | *479.98* | *58.66* | *197.24* | *197.97* | *-0.73* | *50.56* | *50.84* | *-0.28* | *0.63* | *0.69* | *-0.0601* |
| | cubist | 284.55 | 214.58 | 69.97 | 56.40 | 57.64 | -1.24 | 8.64 | 10.22 | -1.58 | 0.90 | 0.94 | -0.0383 |
| | earth | 632.52 | 591.64 | 40.88 | 303.64 | 300.07 | 3.57 | 89.77 | 87.08 | 2.70 | 0.52 | 0.57 | -0.0451 |
| | gbm | 294.33 | 249.05 | 45.28 | 113.54 | 111.77 | 1.77 | 29.22 | 29.88 | -0.66 | 0.90 | 0.92 | -0.0253 |
| | glmnet | 706.97 | 679.03 | 27.93 | 304.67 | 308.15 | -3.48 | 76.78 | 76.92 | -0.15 | 0.40 | 0.43 | -0.0270 |
| | knn | 481.43 | 341.49 | 139.93 | 85.48 | 89.40 | -3.91 | 13.17 | 16.47 | -3.30 | 0.72 | 0.86 | -0.1331 |
| | lm | 706.93 | 677.02 | 29.92 | 305.05 | 308.58 | -3.53 | 77.00 | 77.22 | -0.22 | 0.40 | 0.43 | -0.0303 |
| | ranger | 538.22 | 487.25 | 50.97 | 173.04 | 183.07 | -10.03 | 40.55 | 43.66 | -3.10 | 0.69 | 0.75 | -0.0558 |
| | rf | 541.91 | 475.44 | 66.47 | 175.35 | 181.35 | -6.00 | 41.66 | 43.25 | -1.59 | 0.68 | 0.76 | -0.0750 |
| | rpart | 771.26 | 664.01 | 107.25 | 386.76 | 366.46 | 20.31 | 129.38 | 123.63 | 5.75 | 0.29 | 0.45 | -0.1665 |
| | svmRadial | 687.90 | 661.27 | 26.63 | 190.71 | 198.53 | -7.82 | 34.42 | 34.74 | -0.32 | 0.46 | 0.50 | -0.0434 |
| | xgbTree | 279.00 | 239.00 | 40.01 | 74.95 | 72.67 | 2.27 | 15.56 | 16.16 | -0.61 | 0.91 | 0.93 | -0.0211 |
| *Total* | | 516.65 | 493.36 | 23.29 | 197.12 | 196.61 | 0.51 | 57.26 | 43.66 | 13.60 | 0.65 | 0.68 | -0.0321 |

## Machine Learning Model Selection

Although polymer-specific models can achieve strong local fits, they require large, well-balanced data sets for each material and often generalize poorly to under-represented polymers. Here, a global modeling strategy leverages shared process-structure relationships in electrospinning through routinely reported, experimentally controllable variables (e.g., concentration, voltage, flow rate, tip-to-collector distance, solvent ratios, and ambient conditions), while polymer and solvent identity are encoded explicitly as categorical predictors. Balanced sampling mitigates dominance effects from over-represented polymers and ensures minority materials contribute meaningfully to model training. Polymer-wise inverse performance remains heterogeneous, consistent with genuine differences in process robustness and data coverage rather than systematic model bias.

Table 3 compares the predictive performance of 11 machine-learning algorithms across the three sampling strategies (Balanced Sobol+D-optimal sampling, D-Optimal+Sobol, and Random). Although SpinCastML allows cross-validation k-folds, the main benchmarking results presented in Table 3 correspond to the selected 5-folds configuration used for final model comparison, additional results for 3 and 10-folds can be found in Supporting Information. Preprocessing before model training, features and type of learners are described in the Methods Section.

To evaluate model robustness beyond predictive accuracy alone, model selection was based not only on RMSE, MAE, MAPE, and $R^2$, but also on the agreement between cross-validation and independent test-set performance. This comparison provides an assessment of generalization stability and helps identify potential overfitting or sampling-induced bias. Models exhibiting small discrepancies between cross-validation and test metrics were considered more reliable than models achieving high predictive performance in only a single validation scenario. Consequently, the final model selection reflected a balance between predictive accuracy and consistency across independent validation procedures.

Across metrics (RMSE, MAE, MAPE, $R^2$), the Balanced sampling setup shows the smallest discrepancies between test and cross-validation (mean $\Delta RMSE = -5.17$; $\Delta R^2 = -0.0128$), indicating minimal overfitting and strong generalization. This aligns with evidence that balanced or stratified sampling mitigates dominance of overrepresented classes and improves predictive stability in multisource experimental data sets.[35] Balanced sampling also ensures full representation of minority polymers and reduces redundancy in majority classes, consistent with optimal-design literature validating hybrid D-optimal/Sobol strategies for chemical engineering.[41]

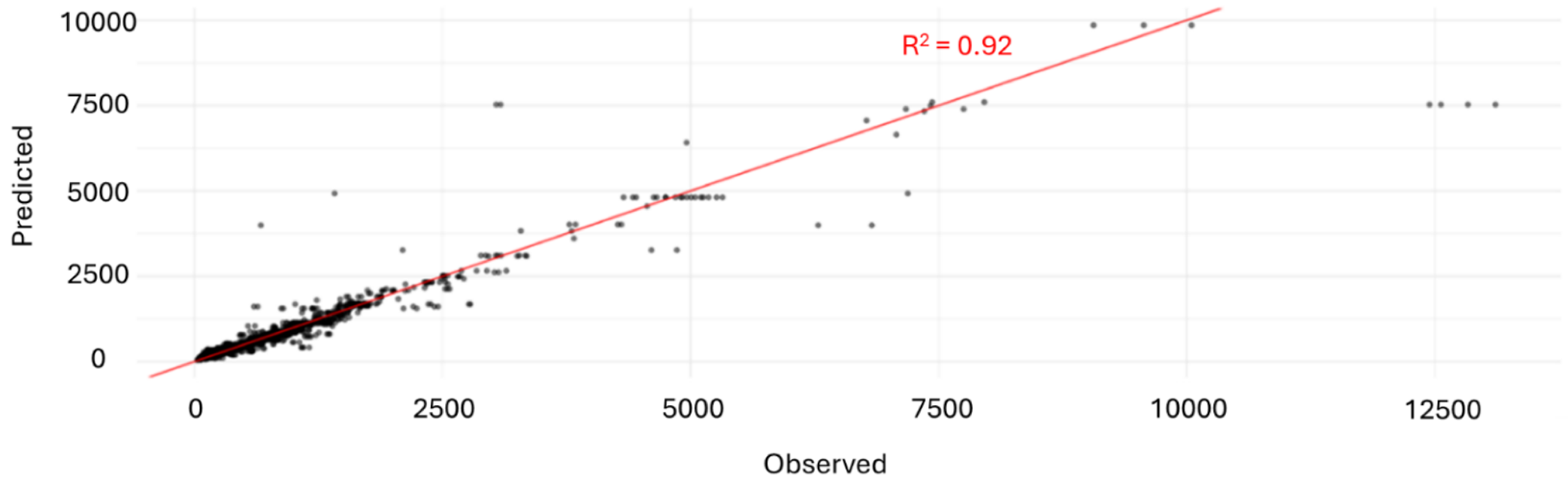

**Figure 2.** "Predicted vs Observed" scatter plot of Cubist model and balanced sampling.

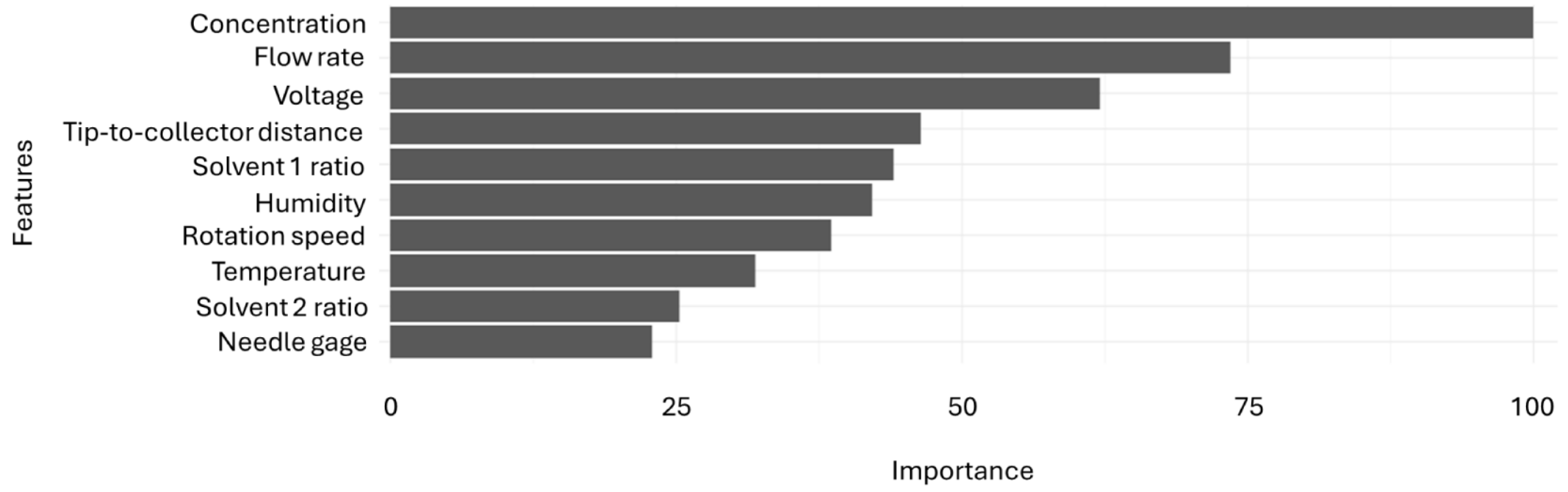

**Figure 3.** Normalized feature importance in electrospun fiber diameter of Cubist model and balanced sampling.

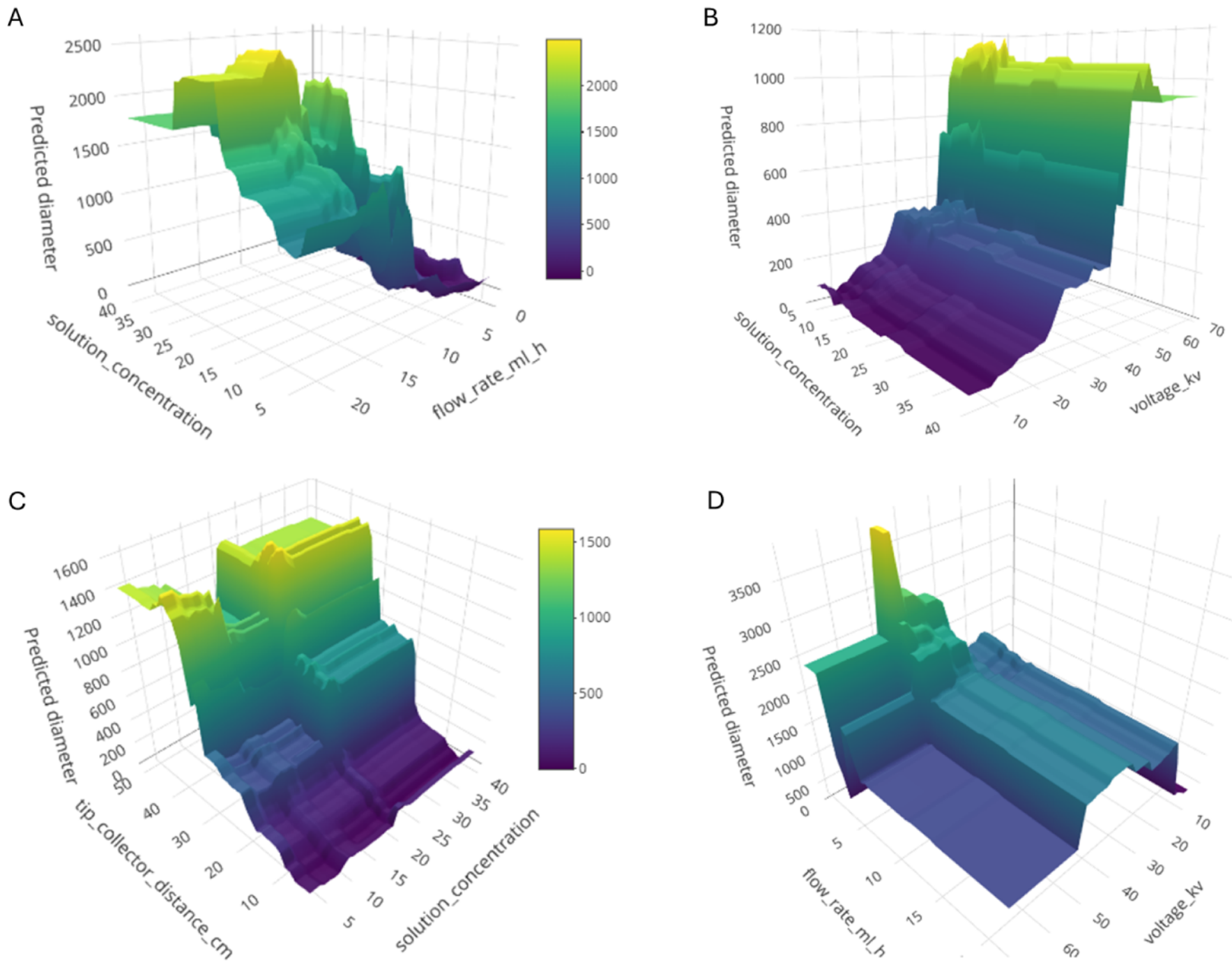

**Figure 4.** 3D response-surface plots with predicted fiber diameter (nm) as a function of solution concentration (%) and A) flow rate (ml/h), B) voltage (kV), C) tip-to-collector distance (cm), and D) flow rate (ml/h) and voltage (kV).

Across sampling strategies, Cubist is consistently among the best-performing models (Balanced RMSE = 260 nm; $R^2 = 0.92$), closely followed by GBM and XGBTree, confirming the strong performance of rule-based and gradient-boosting learners in nonlinear materials-property prediction. This finding was also reported in several studies[15,27,42] with values of $R^2$ of 0.93−0.989.

Models such as glmnet and LM show RMSE values >690 nm and $R^2 \approx 0.40-0.43$, reflecting their inability to capture the strong nonlinearities intrinsic to electrospinning (concentration−viscosity−diameter relationships, field-dependent jet stretching). This echoes prior studies demonstrating that linear models underperform against nonlinear learners.[5,26,29,43]

Random Forest and Ranger show good $R^2$ (0.68−0.76) but large ΔRMSE values under Random and D-Optimal+Sobol

sampling (up to +66), suggesting sensitivity to sample imbalance and redundancy, which is consistent with reports that tree-based ensembles can be affected by highly skewed training distributions.[35,44]

D-Optimal+Sobol achieves broad coverage of the experimental domain (lower MAPE for some models) but shows elevated ΔRMSE and $\Delta R^2$ values (mean ΔRMSE = +57.36; $\Delta R^2 = -0.0236$). This indicates that although the sampling is statistically efficient, it introduces greater divergence between cross-validation (CV) and test sets, an expected behavior when exploring extrapolative regions of the design space.

The Random strategy exhibits the largest ΔRMSE (mean +58.66) and $\Delta R^2$ (−0.0601), highlighting strong overfitting tendencies and instability, especially for algorithms sensitive to uneven class distributions such as kNN (ΔRMSE = +139.93) and rpart (ΔRMSE = +107.25). This supports the view that naïve random sampling is inadequate in imbalanced, multi-polymer data sets.[35]

Overall, the combination of Balanced sampling (polymer-balanced Sobol and D-optimal sampling) and Cubist delivers the strongest and most stable performance (RMSE = 260 nm; MAE = 50 nm; $R^2 = 0.92$; small CV-Test discrepancies), making it the most reliable modeling choice in this study. It reflects the dual advantage of (i) high-capacity nonlinear learning and (ii) balanced sampling that preserves polymer diversity while controlling dominance effects. Figure 2 shows the "Predicted vs Observed" scatter plot for this strategy.

## Interpretability

To gain insight into the relative influence of the processing and formulation variables on fiber diameter, normalized feature-importance rankings and corresponding 3D response surfaces are presented in Figures 3 and 4. These analyses provide interpretable summaries of the relationships learned by the machine-learning model and reveal a hierarchy that is both statistically robust and broadly consistent with established electrospinning principles. Because the framework was trained on a heterogeneous data set comprising multiple polymers, solvent systems, and processing conditions, the observed relationships should be interpreted as data set-level trends rather than universal mechanistic laws. Accordingly, the normalized feature-importance rankings and response surfaces complement existing electrospinning theory by highlighting influential variables and interactions within the experimental domain represented by the available data.

Feature-importance values are reported on a normalized relative scale, with the top-ranked predictor scaled to 100 and the remaining predictors expressed relative to this maximum; therefore, the value of 100 for concentration should not be interpreted as an absolute contribution or percentage of explained variance.

Within the assembled data set, concentration emerges as the most influential feature in the model (Figure 3), in agreement with previous studies[5,15,21] and produces the steepest, most systematic gradients in all response surfaces, reflecting its central role in controlling chain entanglement, viscoelastic stress, and the transition from beaded jets to stable fiber formation, as reported in the literature.[45,46] The response surfaces show that increasing concentration consistently shifts the system into a more stable electrospinning regime, leading to larger and more uniform fibers, mirroring its top-ranked importance in the model. Flow rate and voltage appear next in the importance ranking and their surfaces illustrate how these parameters modulate the mass flux and electrostatic stretching forces acting on the jet. Elevated flow rates increase volumetric throughput and reduce jet residence time, often producing thicker or more irregular fibers, as reported in prior studies.[46] Similarly, voltage influences the balance between Coulombic stretching and viscoelastic resistance, with higher fields promoting jet elongation, a behavior clearly reflected in the predicted diameter gradients and documented in electrospinning theory.[47] The tip-to-collector distance and solvent ratios also show appreciable influence, consistent with their roles in defining the available flight time for whipping instabilities to develop and for solvent evaporation to occur; insufficient distance or slow evaporation rates are known to induce wet fibers or beads.[48] Humidity and temperature, although less influential, align with established sensitivity of fiber morphology to environmental conditions that alter solvent evaporation kinetics and charge dissipation. Lower-ranked variables such as rotation speed and needle gauge still exhibit localized effects,[21] reflecting their conditional influence on postdrawing and initial jet geometry, respectively, consistent with previous observations that their impact becomes pronounced only under specific combinations of viscosity, electric field strength, and throughput.[45]

The concentration versus voltage surface (Figure 4B) reveals smoother, more moderate gradients, consistent with voltage's secondary importance. Although increased electric fields are often associated with enhanced Coulombic stretching and reduced fiber diameter, as described in classical jet-instability studies,[47,49−51] the response surface reflects interactions learned across multiple polymers, solvent systems, and concentration regimes. Consequently, the observed trends should be interpreted as conditional relationships within the experimental domain represented by the data set rather than universal electrospinning behavior. Similarly, the concentration versus tip-to-collector-distance surface (Figure 4C) indicates that the influence of distance depends strongly on concentration and other interacting variables. This observation is consistent with the established role of flight time in governing whipping, jet thinning, and solvent evaporation, where insufficient distance may lead to wet or beaded fibers and excessive distance can produce broader, drier jets.[48,52] The flow-rate versus voltage surface (Figure 4D) further highlights the nonlinear interactions captured by the model. Regions associated with larger predicted diameters should not be interpreted as the isolated effect of voltage or flow rate alone, but rather as the combined influence of formulation and processing variables represented within the heterogeneous data set. Such behavior is consistent with recent quantitative models showing that fiber diameter emerges from complex interactions among electrospinning parameters rather than from the independent action of any single variable.[32]

Additional interpretation graphs such as decision trees and heatmaps importance are presented in Supporting Information.

Overall, the ML-derived hierarchy and the surface topographies converge on a physically coherent picture in which concentration sets the primary electrohydrodynamic regime, while flow rate, voltage, and distance provide secondary, regime-dependent control over jet stability and fiber morphology.

## Inverse Monte Carlo Simulations

Inverse electrospinning design does not admit a single-valued or smooth inverse mapping from target fiber morphology to processing parameters. Small perturbations in solution composition, environmental conditions, or operating settings can trigger regime shifts, such that statistically similar fiber diameters arise from multiple, often disconnected regions of the design space governed by nonlinear electrohydrodynamic effects, solvent evaporation kinetics, and jet instabilities.[51] Deterministic inverse strategies or optimization-based approaches therefore risk collapsing this inherently multimodal solution space toward narrow, fragile optima that lack robustness to experimental variability.[53] Inverse Monte Carlo (IMC) addresses this challenge by reframing inverse electrospinning as a probabilistic inference problem, in which candidate configurations are sampled and evaluated at the distribution level.[54] This formulation preserves solution multiplicity, quantifies uncertainty through success probabilities, and naturally incorporates explicit chemical and operational constraints, enabling inverse design toward polymer-specific diameter architectures and realistic fiber distributions rather than single averaged outcomes.

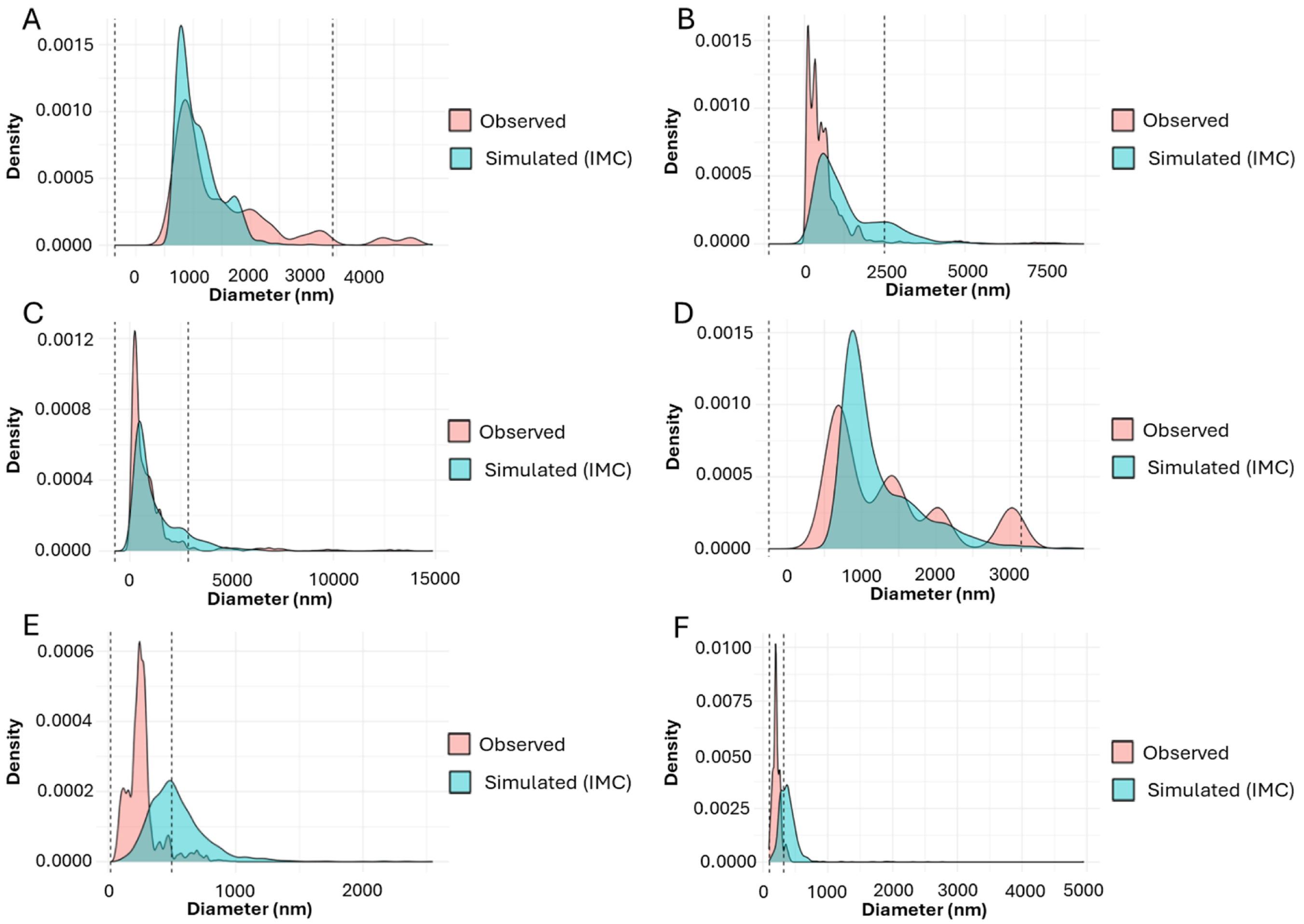

**Figure 5.** Fiber diameter distributions from observed data and predicted distribution (simulated with IMC with balanced sampling and Optimization mode) A) PS, B) PVDF, C) PVP, D) PDLLA, E) PVA and F) PAN.

Analyzing the performance of the IMC framework across polymers, sampling strategies (Random, Sobol+D-optimal, and Balanced), and IMC modes (Experimental and Optimized), we observed that across all sampling strategies, Experimental IMC closely reproduces the empirical fiber diameter statistics. For example, under Random sampling, CA shows a predicted mean of 504 nm and standard deviation of 352 nm, in excellent agreement with the observed mean of 509 nm and standard deviation of 360 nm. Similar agreement is observed for PAN (predicted mean of 208 nm vs observed mean of 209 nm), PCL (361 nm vs 362 nm), and PET (406 nm vs 405 nm), demonstrating that Experimental IMC preserves both central tendency and dispersion across polymers. This behavior is consistent across Sobol+D-optimal and Balanced sampling, confirming that Experimental IMC provides a faithful distribution-level baseline independent of sampling strategy.

In contrast, Optimization mode exhibits a strongly polymer-dependent response. For some polymers, Optimization mode improves or maintains inverse performance. A notable example is PS: under Random sampling, Optimization reduces the predictive standard deviation from 880 to 390 nm while increasing the success probability from 87.4% to 98.8%. Similar improvements are observed under Sobol+D-optimal sampling, where the success probability increases from 94.9% to 99.9%. These results indicate that, for certain polymers, constraining the inverse search toward favorable operating regions leads to more concentrated predictions and higher likelihood of meeting the target diameter.

By contrast, several polymers show a reduced performance under Optimization mode. For PVA, Optimization under Random sampling shifts the predicted mean from 258 to 630 nm and increases the predictive standard deviation from 118 to 426 nm, causing the success probability to drop sharply from 72.5% to 29.1%. A similar pattern is observed for PAN, where Optimization mode increases the predicted mean from 208 to 361 nm and reduces the success probability from 63.6% to 31.8%.

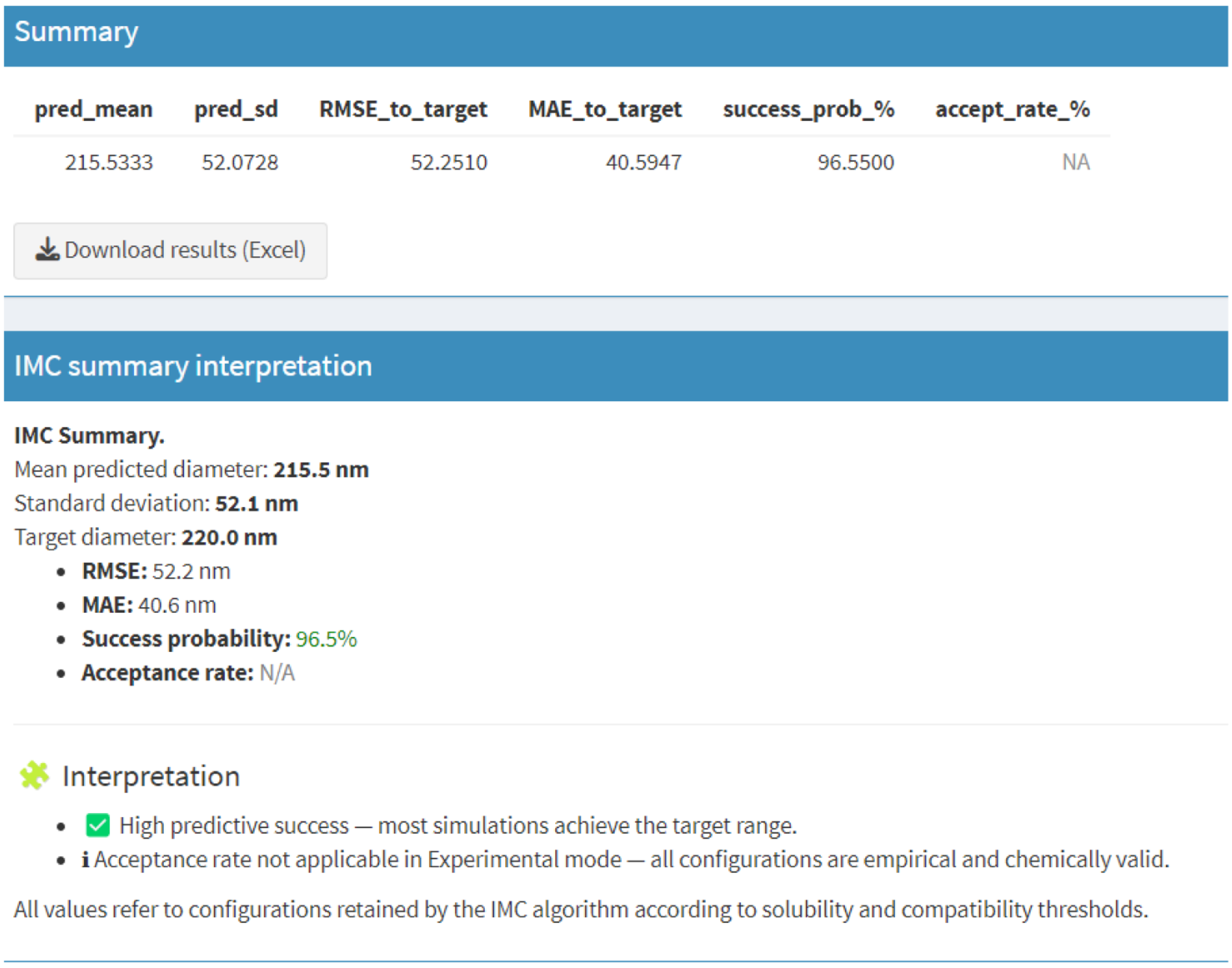

Summary

| pred_mean | pred_sd | RMSE_to_target | MAE_to_target | success_prob_% | accept_rate_% |
|---|---|---|---|---|---|
| 215.5333 | 52.0728 | 52.2510 | 40.5947 | 96.5500 | NA |

Download results (Excel)

IMC summary interpretation

**IMC Summary.**
Mean predicted diameter: **215.5 nm**
Standard deviation: **52.1 nm**
Target diameter: **220.0 nm**

- **RMSE:** 52.2 nm
- **MAE:** 40.6 nm
- **Success probability:** 96.5%
- **Acceptance rate:** N/A

Interpretation

- ✅ High predictive success — most simulations achieve the target range.
- ℹ Acceptance rate not applicable in Experimental mode — all configurations are empirical and chemically valid.

All values refer to configurations retained by the IMC algorithm according to solubility and compatibility thresholds.

**Figure 6.** Screenshot of IMC summary for PAN in Experimental mode.

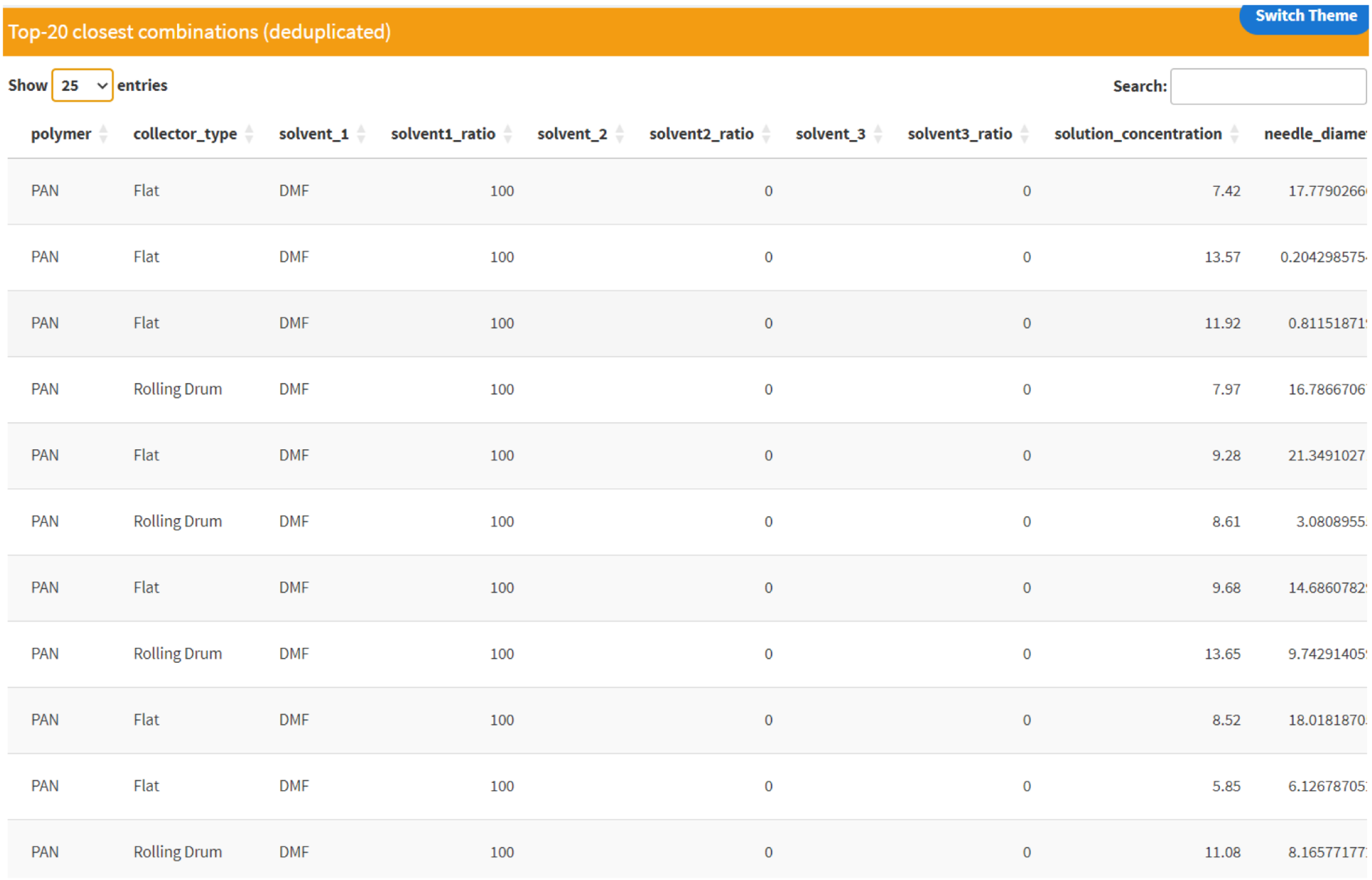

Top-20 closest combinations (deduplicated) Switch Theme

Show 25 entries Search:

| polymer | collector_type | solvent_1 | solvent1_ratio | solvent_2 | solvent2_ratio | solvent_3 | solvent3_ratio | solution_concentration | needle_diame |
|---|---|---|---|---|---|---|---|---|---|
| PAN | Flat | DMF | 100 | | 0 | | 0 | 7.42 | 17.7790266 |
| PAN | Flat | DMF | 100 | | 0 | | 0 | 13.57 | 0.204298575 |
| PAN | Flat | DMF | 100 | | 0 | | 0 | 11.92 | 0.81151871 |
| PAN | Rolling Drum | DMF | 100 | | 0 | | 0 | 7.97 | 16.7866706 |
| PAN | Flat | DMF | 100 | | 0 | | 0 | 9.28 | 21.3491027 |
| PAN | Rolling Drum | DMF | 100 | | 0 | | 0 | 8.61 | 3.0808955 |
| PAN | Flat | DMF | 100 | | 0 | | 0 | 9.68 | 14.6860782 |
| PAN | Rolling Drum | DMF | 100 | | 0 | | 0 | 13.65 | 9.74291405 |
| PAN | Flat | DMF | 100 | | 0 | | 0 | 8.52 | 18.0181870 |
| PAN | Flat | DMF | 100 | | 0 | | 0 | 5.85 | 6.12678705 |
| PAN | Rolling Drum | DMF | 100 | | 0 | | 0 | 11.08 | 8.16577177 |

**Figure 7.** Screenshot of the top-20 list of electrospinning setup predicted for PAN in Experimental mode.

Importantly, these polymer-dependent trends are consistent across all three sampling strategies. While Balanced (polymer-balanced Sobol and D-optimal sampling) and Sobol+D-optimal sampling slightly improve the agreement between predicted and observed success probabilities in Experimental mode (e.g., PVDF success probability ≈ 92.5−92.9%), they do not prevent the deterioration observed under Optimization mode (e.g., PVDF success probability ≈ 59.8−63.8). This demonstrates that inverse predictability is governed primarily by polymer-process behavior rather than by data imbalance alone.

A key observation is that mean agreement alone is insufficient to assess inverse design performance. For instance, PVDF maintains close agreement between predicted and observed mean diameters under both modes (predicted mean ≈ 670−690 nm vs observed mean ≈ 669−690 nm), yet Optimization dramatically increases predictive dispersion (predicted standard deviation rising above 1000 nm) and reduces success probability from ∼93% to ∼64%. This decoupling between mean accuracy and probabilistic success highlights the importance of evaluating inverse design outcomes using variance-sensitive metrics.

Overall, these results demonstrate that SpinCastML and its IMC engine provide a powerful and flexible framework for inverse electrospinning design. Experimental IMC reliably reproduces known behavior, while Optimization mode can either sharpen inverse solutions or reveal the breadth of feasible operating conditions, depending on the polymer system. Rather than prescribing a single optimal recipe, IMC offers a quantitative map of achievable performance, enabling users to make informed, polymer-specific design decisions grounded in both predictive accuracy and operational robustness. Data supporting these results can be found in Supporting Information.

Figure 5 shows the comparison between fiber diameter distributions from observed data and predicted distribution calculated with polymer-balanced Sobol+D-optimal sampling and Optimization mode for six different polymers.

Screenshots of the IMC summary and the top-20 list of electrospinning setup predicted by SpinCastML for PAN in Experimental and Optimization mode respectively can be found in Figures 6 and 7.

## SpinCastML User Guide

SpinCastML is an interactive, reproducible, standalone framework that combines machine learning with inverse Monte Carlo to identify feasible electrospinning conditions and solvent formulations that achieve targeted, polymer-specific nanofiber diameter distributions. SpinCastML is released in an open-source repository with versioned releases, including the full application code, the curated data set schema, and the chemical feasibility resources required to reproduce all analyses reported here. The software developed in this study is released under the GNU GPL 3.0 license and is openly available as a citable archive at Zenodo (10.5281/zenodo.18557989),[55] including code, input files, instructions and license.

This SpinCastML user guide ensures reproducibility, transparency, and community reuse of the proposed methodology rather than to emphasize interface design. The application operationalizes the modeling, sampling, and inverse Monte Carlo framework described above, enabling consistent application to new data sets without requiring specialist machine-learning expertise. Interface details are included for completeness and reproducibility, while the primary scientific contribution of this work resides in the distribution-aware modeling strategy, chemically constrained inverse design, and probabilistic interpretation of electrospinning outcomes.

SpinCastML is a self-contained Shiny app. Users should install the SpinCastML in their desired folder and start the launcher (run_app.bat) or run shiny:run_app(run_app.R) in R. The app links to the R code SpinCastML.R and opens in a web browser with a left control sidebar and tabbed workspace.

In the Data & Preprocessing module, users may either upload an experimental database (.xlsx) or use the data set supplied in Templates/DATA_BASE; when the supplied data set is selected, no file upload is required. Column headers and solvent names are harmonized automatically. A solubility status message (OK/COND/NO counts) confirms chemical constraints were loaded.

For model training, user must set a random sample size, a hybrid Sobol−D-optimal sampling, or select a polymer-balanced Sobol and D-optimal sampling scheme, test share (10−40%, default 30%), and CV folds (3/5/10), and select candidate algorithms and click "Train models". The app fits a preprocessing recipe on train only, performs a desired train-test split, and runs k-fold CV.

In Metrics tab, the best model is chosen by minimum test RMSE. A metrics table reports RMSE, MAE, MAPE, and $R^2$ for CV and hold-out. Diagnostic plots include Observed vs Predicted, Residuals vs Predicted, and residual QQ-plot with clear indications of the results. Users can export/import the fitted model and recipe as a single.RDS file.

After the models are validated, the Inverse Monte Carlo can be performed to obtain the electrospinning settings and solution design to get the desired morphology of the electrospun scaffolds. To perform Inverse Monte Carlo, users should choose Experimental (sampled from the empirical distribution) or Optimization (synthetic configurations within observed ranges). Specify polymer, solubility strictness (Strict/Balanced/Lax), optional allowance for NO pairs, target diameter and tolerance, and number of simulations. Outputs report mean/SD of predictions, error to target (RMSE/MAE), success probability, and (in optimization mode only) acceptance rate. A density plot and a deduplicated Top-20 table (chemically viable combinations) are provided; results can be exported to Excel.

The interpretability tab provides global variable importance, a two-variable heat map, and a 3-D prediction surface. Other results collate data set/model diagnostics, regression coefficients (lm/glmnet), a surrogate decision tree, Cubist rules (when applicable), and a one-click PDF report bundling active metrics, diagnostics, and IMC outputs.

Recommended workflow. Install SpinCastML, run run_app.bat, load data (own one or our Data_Base file) and verify solubility; train multiple algorithms; select the best model from Metrics; save the.RDS; run IMC in Experimental then Optimization modes to benchmark feasibility and explore admissible space; export Top-20 and/or full report for laboratory review. Reproducibility is ensured because predictions reuse the saved preprocessing pipeline and model.

A typical application of SpinCastML involves specifying a target fiber-diameter distribution and tolerance for a selected polymer, followed by inverse exploration of chemically feasible processing conditions. Experimental-mode IMC first evaluates whether historical data support the desired outcome, while Optimization mode probes admissible regions of the formulation and operating space. Rather than returning a single optimized recipe, the framework produces a ranked set of feasible configurations with quantified success probabilities, allowing users to balance robustness, solvent choice, and operational flexibility. This workflow demonstrates how inverse electrospinning design is transformed from deterministic prescription into probabilistic decision support, reducing trial-and-error experimentation and improving reproducibility.

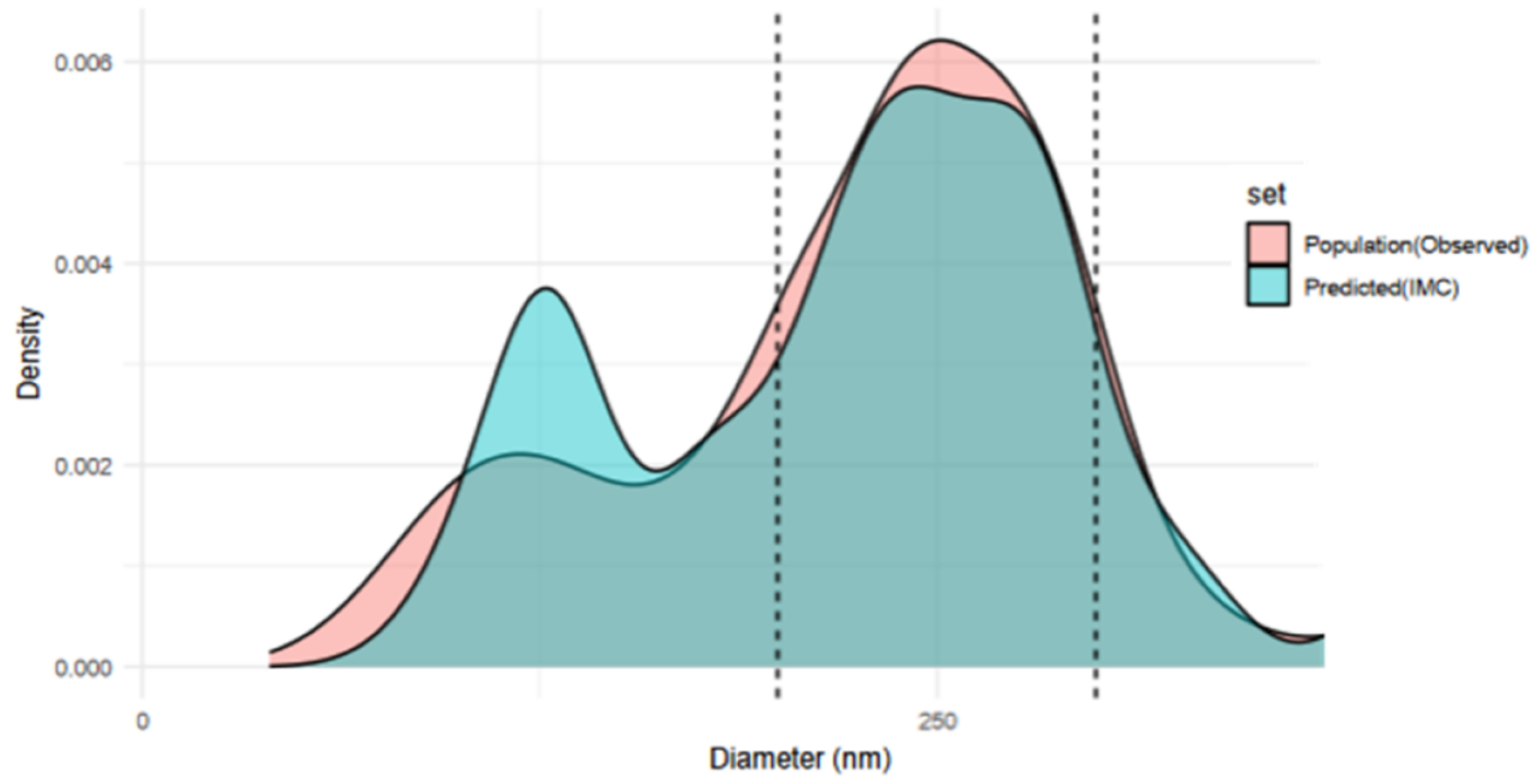

**Figure 8.** Fiber diameter distributions from observed data and predicted distribution for PVA.

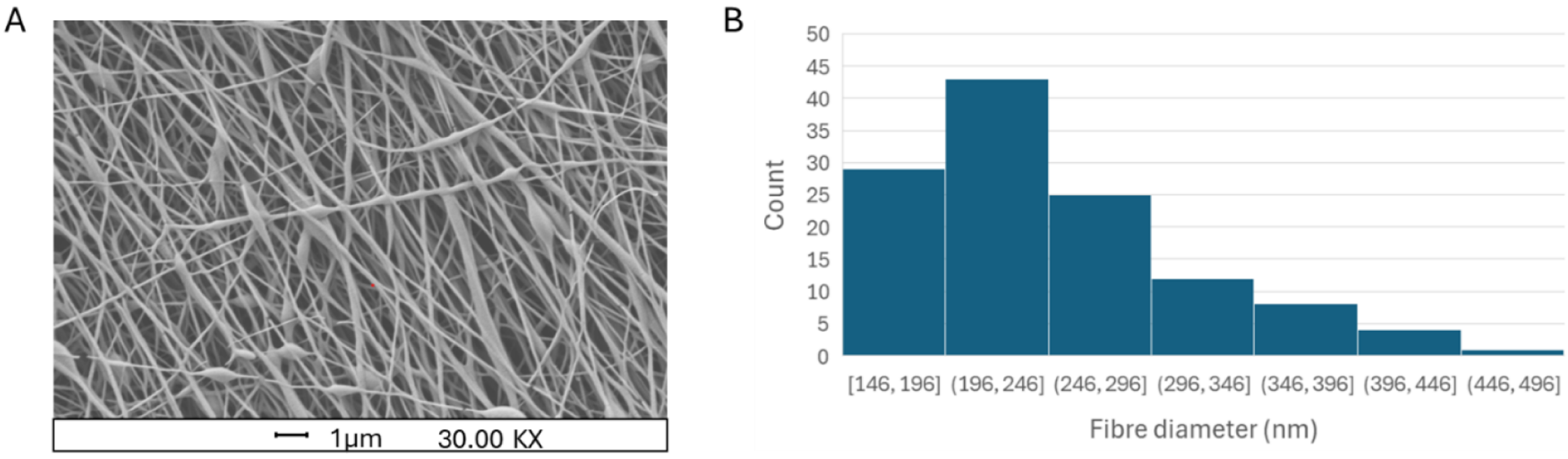

**Figure 9.** PVA prospective case study A) SEM produced from IMC-derived formulation B) Experimental fiber-diameter distribution.

Note that, when users upload independently curated data sets, SpinCastML applies the same preprocessing, sampling, model-training, validation, and reporting workflow used for the preloaded database, provided that the uploaded file contains the required fields in the same schema. In this mode, model behavior depends on the completeness, size, balance, and experimental coverage of the user-provided data. Data sets with sufficient representation across the relevant polymer, solvent, and processing-variable ranges are expected to support more reliable data set-specific models, whereas small, incomplete, or highly imbalanced data sets may yield reduced predictive performance. For this reason, users should interpret predictions and inverse-design outputs together with the train-test, cross-validation, and diagnostic metrics generated by SpinCastML.

## Prospective Case Study

To demonstrate the practical feasibility of SpinCastML, the trained Cubist model was applied to an inverse-design case study using PVA as the target polymer. The design objective was to identify electrospinning conditions capable of producing fibers with a target diameter of 250 nm and a tolerance of ±50 nm, corresponding to an acceptable range of 200−300 nm.

The IMC module generated 20 chemically feasible PVA formulation−process families, reported in the Supporting Information. All candidate families were based on PVA dissolved in water, used a flat collector, and were classified with an OK solubility flag, confirming that the suggested solutions satisfied the chemical-feasibility rules implemented in SpinCastML. The predicted diameters of the 20 families ranged from 240.27 to 259.73 nm, with absolute errors between 0.47 and 9.73 nm relative to the 250 nm target. This narrow prediction range indicates that the IMC module was able to identify multiple experimentally plausible routes close to the desired fiber diameter, rather than a single isolated solution.

From these 20 candidate families, one condition was selected for experimental validation. The selected condition corresponded to a 12% w/w PVA solution in water, electrospun using a 22 G needle, 25 kV, 0.3 mL $h^{-1}$ flow rate, 10 cm tip-to-collector distance, 20 °C, and 45% relative humidity. Although this candidate was not the closest prediction numerically, it was selected because it combined a low prediction error with the highest feasible PVA concentration among the candidate families. This decision was made to improve productivity, since increasing the polymer concentration increases the amount of fiber-forming material delivered per unit volume of solution and may reduce solvent demand for producing a given scaffold mass.

The IMC-predicted distribution showed strong agreement with the observed PVA population distribution. For the IMC output, the predicted distribution contained n = 2000 simulated fibers, with a mean diameter of 256.94 nm and a standard deviation of 121.07 nm. The observed PVA population contained n = 967 fibers, with a mean diameter of 258.85 nm and a standard deviation of 125.72 nm (Figure 8). The difference between the sampled and population means showed not statistically significant at the 95% confidence level ($p = 0.17$). The close overlap between the predicted and observed density curves supports the ability of the IMC workflow to reproduce the central tendency and dispersion of experimentally reported PVA fiber diameters.

The selected condition was then tested experimentally using electrospinning. SEM analysis confirmed the formation of a fibrous PVA scaffold, demonstrating that the IMC-derived formulation and processing parameters were experimentally feasible. The SEM showed a nonwoven network of randomly oriented fibers, with a substantial proportion of fibers in the nanoscale range. The experimentally measured fiber-diameter distribution should be compared with the predefined target range of 200−300 nm. The measured mean and standard deviation diameter were 243.59 ± 70.39 nm, indicating that the experimental scaffold fell within the target window predicted by SpinCastML (Figure 9).

This prospective case study strengthens the validation of the framework in two important ways. First, it demonstrates that the IMC output can be translated into a realistic laboratory condition after applying practical decision rules such as parameter rounding, equipment compatibility, and productivity optimization. Second, it provides experimental evidence that SpinCastML can reduce trial-and-error optimization by narrowing the design space to a small number of feasible candidate families.

### Limitations of the Study

Several curated physicochemical and rheological descriptors known to influence fiber formation are not yet leveraged in SpinCastML. Incorporation of variables such as steady-state and zero-shear viscosity, electrical conductivity, surface tension, polymer molecular weight and polydispersity, solvent volatility, and temperature-dependent properties could enhance the model's capacity to reflect the governing transport and instability physics. Likewise, inclusion of derived or dimensionless quantities (e.g., Ohnesorge and Reynold's numbers, Deborah/Weissenberg metrics, and Hansen/Flory-Huggin's interaction parameters) may provide a more mechanistic signal than raw operating settings. Collectively, these enhancements are expected to improve model robustness, interpretability, and predictive utility for electrospinning optimization. Future work will therefore seek to incorporate these predictors, alongside expanding the database to polymer blends and additives (such as surfactants, salts, nanoparticles), to further enhance model performance and advance understanding of the key factors influencing fiber formation.

The SEM images performed during the case study revealed an important limitation of the current model. While SpinCastML successfully targets fiber diameter, the present version does not explicitly predict secondary morphological features such as bead formation, fiber uniformity, surface texture, interfiber fusion, or scaffold porosity. The observed morphology suggests that, although the selected condition was able to generate fibers within the desired diameter regime, additional optimization may be required to minimize bead-on-string features and improve morphological homogeneity. Future versions of the tool should therefore incorporate additional output descriptors, including bead density, fiber continuity, alignment, and interfiber spacing, to move beyond diameter prediction toward more complete scaffold morphology design.

In addition, although this study focused on sampling-based acceleration, alternative scalability strategies may further improve future versions of SpinCastML. Kernel approximation methods such as the Nyström approximation, random Fourier features, approximate nearest-neighbor search, dimensionality reduction, and parallel model training could reduce computational cost for specific learning algorithms or larger feature spaces. In the present work, sampling-based strategies were prioritized because they simultaneously reduced data set size, improved coverage of the experimental domain, and addressed polymer imbalance, whereas kernel approximations primarily accelerate model fitting without directly correcting representational bias. Future implementations may combine balanced sampling with algorithm-level acceleration methods to further improve scalability.

## CONCLUSIONS

This study establishes SpinCastML as a distribution-aware, chemically constrained framework for inverse electrospinning design, addressing fundamental limitations of conventional forward-prediction and trial-and-error approaches. By explicitly modeling fiber-diameter distributions rather than single-point estimates, SpinCastML captures the intrinsic variability and regime-dependent behavior that define electrospinning outcomes across polymers, solvents, and operating conditions.

A central contribution of this work is the formulation of inverse electrospinning as a probabilistic inference problem. The proposed Inverse Monte Carlo (IMC) engine acknowledges the nonuniqueness of the inverse mapping between target morphology and processing parameters, returning families of chemically feasible solutions with quantified success probabilities instead of prescribing a single fragile optimum. This distribution-aware inverse perspective enables users to reason about robustness, risk, and operational flexibility, which are critical for reproducible manufacturing in practice.

The integration of polymer−solvent solubility and solvent−solvent compatibility constraints ensures that inverse-design outputs remain experimentally actionable. By embedding chemical feasibility directly into both model training and inverse simulation, SpinCastML aligns data-driven predictions with laboratory reality and supports safer, more sustainable exploration of formulation space. The use of structured, polymer-balanced Sobol and D-optimal sampling further enhances model stability and generalization across heterogeneous, imbalanced experimental data sets.

Beyond methodological advances, SpinCastML operationalizes these concepts in an open, executable application that lowers the barrier to adoption of machine learning and inverse design in electrospinning research. The framework enables researchers to analyze their own data sets, reproduce results transparently, and contribute to a growing community-driven resource, fostering cumulative progress rather than isolated optimization studies.

Importantly, the contribution of this work lies not in incremental improvements in diameter prediction accuracy, but in a conceptual shift: reframing electrospinning design as a distributional, inverse, and chemically constrained problem under realistic experimental variability. This paradigm is particularly relevant for emerging challenges such as blended polymer systems, greener solvent selection, and application-driven nanofiber architectures, where deterministic optimization is insufficient. By unifying distribution-aware modeling, chemical feasibility, and probabilistic inverse design, SpinCastML establishes a new standard for data-driven decision-making in electrospinning and provides a generalizable blueprint for inverse design in complex manufacturing processes.

## METHODS

### Data Collection

A rigorously curated data set was assembled through a systematic survey of Scopus and Google Scholar. Searches combined the term "electrospinning" with widely used polymers, including cellulose acetate (CA), gelatin, polyamide-6 (nylon-6), polyacrylonitrile (PAN), polycaprolactone (PCL), poly(D,L-lactide) (PDLLA), polyether−ether−ketone (PEEK), polyethylene terephthalate (PET), polylactic acid (PLA), poly(methyl methacrylate) (PMMA), polystyrene (PS), polyurethane (PU), poly(vinyl alcohol) (PVA), polyvinylidene fluoride (PVDF), polyvinylpyrrolidone (PVP), and poly-$\gamma$-glutamic acid ($\gamma$-PGA). Eligibility was restricted to fully reproducible experimental articles that (i) used a single polymer and (ii) reported fiber-diameter distributions including histograms or descriptive statistics of fiber-diameter obtained from scanning electron microscope. Two published data sets[56,57] were incorporated, yielding a final corpus of 1,778 experimental studies.

The data set was curated through a structured manual extraction and harmonization workflow. Manual curation was necessary because electrospinning studies report experimental conditions in heterogeneous formats, including different units, solvent nomenclature, concentration notation, collector descriptions, and levels of reporting completeness. For each study, the following fields were extracted where available: document identifier (DOI); polymer; solvents 1, 2, and 3 and their ratios (%); polymer concentration (%); needle diameter; collector type; rotation speed (rpm); applied voltage (kV); flow rate (mL/h); tip-to-collector distance (cm); temperature (°C); relative humidity (%); and the reported distribution of fiber diameters. In aggregate, the 1,778 studies contributed 68,480 fiber-diameter observations.

To reconcile inconsistencies across heterogeneous literature sources, polymer and solvent names were standardized, equivalent terminology was harmonized, and reported units were converted to common formats prior to modeling. Entries were retained only when the experimental conditions could be reliably interpreted from the source article. Records with missing essential identifiers or noninterpretable formulation or processing information were excluded rather than inferred manually. Obvious inconsistencies, duplicate entries, and conflicting records were checked during manual curation to preserve experimentally meaningful variability while minimizing errors arising from heterogeneous reporting.

This data set is preloaded within the SpinCastML application. Alternatively, researchers can use SpinCastML to upload their own curated data sets containing the same experimental parameters described above, enabling model training and evaluation without the need for prior machine-learning knowledge. To encourage transparency, reproducibility, and community cocreation, SpinCastML provides the shared data set structure, application code, and chemical compatibility lists.

## Descriptive Analysis of Data Set

The diameter of the fiber was analyzed and stratified by polymer. Data were imported, cleaned and organized with the libraries readxl, dplyr and tidyr, respectively. For each set (overall diameter and by polymer), mean, standard deviation, median, interquartile range (IQR = Q3−Q1), skewness, and excess kurtosis were computed.

Descriptive analysis, data preprocessing and machine-learning modeling were carried out in R (version 4.3.2) within RStudio (version 2024.04.2). All libraries and functions cited in this study were implemented and executed in this software environment.

## Data Preprocessing

All data preparation was scripted to ensure full reproducibility. The raw data set was ingested and column headers were standardized to a consistent naming scheme, mapping legacy labels to the working schema where needed. Solvent mixture proportions were parsed as numeric, set to 0 when the corresponding solvent field was missing, and normalized within each row so that the three proportions sum to 100%. Records were restricted to experiments with stable fiber formation. The outcome variable (fiber diameter, nm) was coerced to numeric, and rows with nonfinite values were removed. Core categorical fields (polymer, solvent_1 to solvent_3, collector type, and a document identifier doi) were cast as factors, and numeric fields were checked for finiteness.

Two auxiliary resources supported later chemistry-aware constraints. First, a solvent−solvent incompatibility list was loaded, with a conservative, hard-coded fallback used if the external file was unavailable. Second, a polymer−solvent solubility table was read that records a categorical rating (OK (polymer soluble in the solvent), COND (conditional, soluble under formulation conditions such as cosolvents), or NO (polymer not soluble in the solvent)) and, when available, a max_pct value. Here, max_pct denotes the maximum allowable proportion (in percent of the total solvent mixture) for that specific solvent with the given polymer under a COND rating. Both resources were cleaned using a standardized solvent naming routine so that lookups are consistent across sources.

Feature engineering and model-ready preprocessing were implemented with a recipes pipeline fit only on the training split to prevent information leakage, then applied identically to cross-validation folds, the held-out test set, and Monte Carlo simulations. In all cases, the random seed was fixed to ensure reproducibility. The pipeline consists of removal of zero-variance predictors and centering and scaling all numeric predictors. Median imputation is appropriate for this setting because it is robust to skewed or heavy-tailed distributions and does not assume normality of fiber diameters.

## Sampling Strategy

To construct an information-rich yet computationally efficient data set, we implemented a multistage sampling strategy combining random sampling, low-discrepancy (Sobol) designs and D-optimal experimental design, and polymer-balanced stratification.

As a baseline, we used simple random sampling without replacement. Given a full data set with N rows and a desired sample size n, a subset S is drawn such that each row has the same inclusion probability (eq 1) with no row selected more than once.

$$Pr(\mathrm{i} \in S) = \frac{n}{N},\ \mathrm{i} = 1, \ldots, N \tag{1}$$

This procedure preserves the empirical joint distribution of all variables but does not enforce uniform coverage in high-dimensional predictor spaces, producing uneven coverage of multivariate continuous variables (e.g., concentration, voltage, flow rate), and may under-representing minority polymers and under-sample rare polymer−solvent combinations due to their low frequency in the population.

To improve space-filling properties, we generate a Sobol low-discrepancy sequence over the hypercube $[0,1]^{d}$, where d is the number of continuous predictors. A Sobol sequence produces quasi-random points that minimize the discrepancy $D_N$ (eq 2):

$$D_N = \sup_{t \in [0,1]^d} \left| \frac{1}{N} \sum_{i=1}^{N} 1(u_i \le t) - \prod_{j=1}^{d} t_j \right| \quad (2)$$

where $\mathbf{1}$ $(u_i \le t)$ is an indicator function. Small $D_N$ values correspond to near-uniform coverage of the domain, reducing clustering, filling gaps, and improving the robustness of downstream predictive models by limiting extrapolation.[38,39] Each Sobol point is then linearly scaled to the empirical range of each numerical variable, ensuring physically realistic candidate combinations (e.g., feasible voltages, concentrations, temperatures, and solvent ratios). These scaled Sobol points constitute the candidate pool for optimal design selection.

Although Sobol sequences improve geometric coverage, they do not account for predictor correlations. To identify the statistically most informative subset, a D-optimal design[40,58] on the Sobol candidate pool using the optFederov() algorithm was applied. Given a design matrix X with p predictors, the D-optimality criterion maximizes (eq 3):

$$\Phi_D = [det(x^T x)]^{1/P} \quad (3)$$

which minimizes the generalized variance of parameter estimates and increases predictor orthogonality.

This hybrid Sobol+D-optimal strategy is similar to two-phase space-filling and model-based design approaches used in chemical engineering,[41] and provides excellent coverage while maintaining statistical efficiency.

Because polymers differ in representation, unconstrained D-optimality may favor dominant polymers. To ensure equitable representation, we adopt a logarithmic weighting allocation (eq 4):

$$n_p = n \frac{log(1 + f_p)}{\sum_q log(1 + fq)} \quad (4)$$

where fp is the frequency of polymer p in the full data set; fq is the frequency of polymer q, where q indexes all polymers in the data set; np is the number of samples allocated to polymer *p* in the balanced sampling scheme; and n is the total desired sample size after sampling, in our case 10000. This approach softly down-weights majority polymers, guarantees inclusion of minority ones, and preserves chemically relevant diversity.

Within each polymer stratum, the Sobol+D-optimal procedure is rerun independently, producing a balanced yet statistically optimal subset. Categorical variables (polymer identity, solvents 1−3, collector type) were sampled from their empirical distribution, with an optional constraint that polymer−solvent tuples must correspond to combinations previously observed experimentally. This prevents chemically implausible formulations while allowing exploration of new regions of the continuous parameter space. Whereas solvent ratios, which must sum to 100%, are generated via a Dirichlet-style transformation and restricted to experimentally observed polymer−solvent compatibilities. This ensures chemically valid mixtures (ratios summing to 100%) while maintaining variability across candidate formulations. This balanced scheme aligns with structured/weighted sampling literature for heterogeneous scientific data sets[35,59] and ensures that models trained on the resulting data generalize well across polymers with differing physicochemical behaviors.

Overall, this sampling strategy preserves empirical chemistry, improves coverage of the multidimensional design space, limits redundancy from oversampled regions, and yields an information-rich yet computationally manageable data set for machine-learning model development.

## Prediction Models

The supervised learning task was defined on the cleaned electrospinning data set, with measured fiber diameter as the outcome and inputs comprising routinely controlled operating conditions (solution concentration, needle gauge, rotation speed, voltage, flow rate, tip-to-collector distance, temperature, humidity), material descriptors (polymer identity and up to three solvents with their percentages), and selected design factors (e.g., collector type). DOI was retained solely for provenance tracking and auditability and was not used as a predictive feature during model training or inference. Fiber diameter prediction was formulated as a supervised regression problem and implemented in caret using a reproducible training-validation workflow. After preprocessing, up to 68,480 observations were subsampled as described in the sampling strategy subsection. A stratified, user-chosen train-test split was then created (default 70/30).

To prevent information leakage, a recipes pipeline was fit on the training partition and then applied unchanged to both training and test data. The same training-derived encoding and scaling were applied to Sobol candidates prior to the D-optimal selection, ensuring no leakage and guaranteeing column alignment. The pipeline handled novel or unknown factor levels, consolidated rare categories (∼1%), performed one-hot encoding of all factors, imputed numeric predictors using the median, removed zero-variance terms, and centered and scaled numeric features. All transformation parameters were learned strictly on the training partition and applied as-is to the hold-out test set and to any Sobol-generated candidate sets and D-optimal selections, ensuring no leakage and guaranteeing column alignment. Basic sanity checks verified adequate target variability (nonzero standard deviation) and the presence of predictors.

A suite of regression algorithms was evaluated, training only those whose dependencies were available at run time: random forest (rf), ranger, radial-kernel support vector regression (svmRadial), elastic net (glmnet), regression tree (rpart), linear model (lm), k-nearest neighbors (knn), gradient-boosted trees (xgbTree, gbm), multivariate adaptive regression splines (earth), and Cubist. Hyperparameters were explored using compact, model-specific grids where appropriate (e.g., mtry near $\sqrt{p}$ for random forest and ranger; nrounds, max_depth, and eta for xgbTree; n.trees, interaction.depth, shrinkage, and n.minobsinnode for gbm); otherwise, a default tuneLength equal 3 was used. Training employed k-fold cross-validation (3, 5, or 10 folds) as configured by the user. Performance was summarized using RMSE, MAE, MAPE, $R^2$ on both cross-validation predictions and the selected hold-out test set. Model selection prioritized the minimum RMSE on the test set, with the minimum cross-validated RMSE used as a fallback when needed. The selected model, together with its preprocessing object, was serialized to a single R's native single-object binary file format (RDS) bundle, which preserves the exact structure and attributes of the saved object for faithful reloading across sessions and systems, to guarantee identical transformations at inference and for subsequent Inverse Monte Carlo analyses.

Aggregating out-of-fold predictions provides the empirical distribution of modeled fiber diameters used in the interface's histograms and predicted-versus-observed plots, delivering both a calibrated point prediction and an interpretable sense of spread

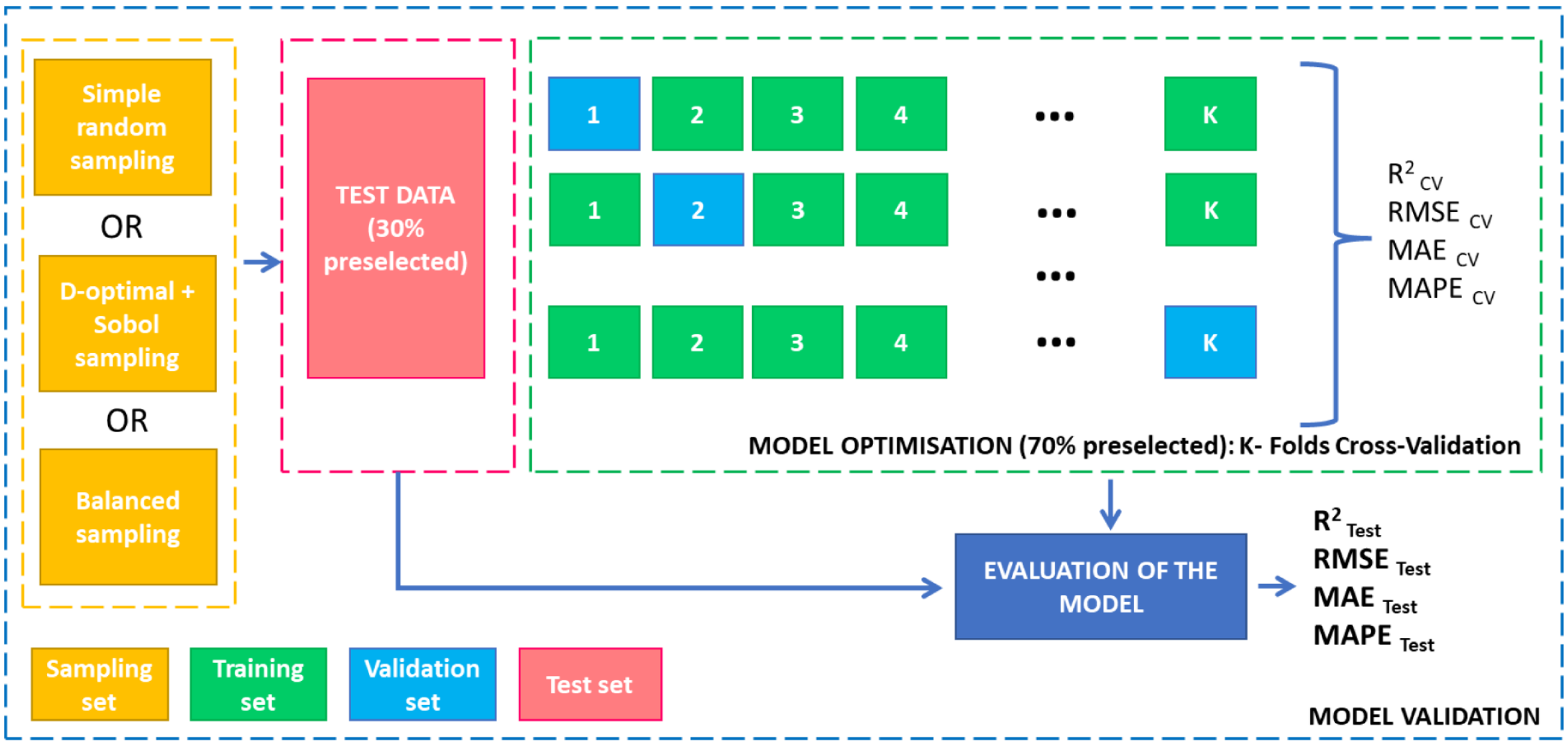

**Figure 10.** Validation of the models.

for decision-making. $R^2$, RMSE, MAE and MAPE were calculated following the eqs 5–8:

$$R^2 = 1 - \frac{\sum_{i=1}^{n} (y_i - \hat{y}_i)^2}{\sum_{i=1}^{n} (y_i - \bar{y})^2} \quad (5)$$

$$RMSE = \sqrt{\frac{1}{n} \sum_{i=1}^{n} (y_i - \hat{y}_i)^2} \quad (6)$$

$$MAE = \frac{1}{n} \sum_{i=1}^{n} |y_i - \hat{y}_i| \quad (7)$$

$$MAPE = \frac{100\%}{n} \sum_{i=1}^{n} \left| \frac{y_i - \hat{y}_i}{y_i} \right| \quad (8)$$

Where $y_i$ is the actual value, $\hat{y}_i$ is the predicted value, $\bar{y}$ is the mean of the actual values, and n is the number of samples.

It is worth noting that model performance was assessed using repeated k-fold cross-validation (k = 3, 5, and 10) combined with an independent hold-out test set, rather than GroupKFold or nested cross-validation, following previous studies.[5,15,31,33] This choice reflects the intended deployment of SpinCastML as a forward-inverse design tool operating over a continuous, high-dimensional process space, rather than as a study-level generalization model. In this work, inverse design is aimed at identifying novel polymer–solvent-process combinations within the learned design space rather than reproducing previously reported experimental groupings. Moreover, future extensions of the framework will explicitly target inverse design of blended polymer systems, for which strict study-level separation imposed by GroupKFold or nested cross-validation would be misaligned with the intended predictive and generative use case. The combined k-fold and hold-out strategy therefore provides a balanced and realistic assessment of predictive performance while preserving the dense coverage of the continuous design space required for reliable inverse Monte Carlo sampling and future extension to blended polymer systems.

Figure 10 shows the validation process of the machine learning models.

## Interpretability

Model behavior was examined through a set of post hoc interpretability analyses applied consistently to the preprocessed features. Global importance scores were computed with caret:varImp. These scores were used to identify the most influential predictors and to guide subsequent response-surface visualizations. To characterize pairwise effects, a heatmap of predicted fiber diameter was generated by varying two operational variables across their observed ranges while holding all remaining inputs at their training medians (for numeric features) or modes (for factors). A complementary 3D surface was produced using the same conditioning strategy, enabling inspection of ridges, valleys, and nonlinear interactions in the prediction landscape.

For structure-aware explanations, a shallow surrogate regression tree (rpart) was fit to the model's training predictions to expose human-readable split rules and interactions, thereby approximating the black-box decision logic. Where applicable, intrinsic model summaries were reported: linear and elastic-net fits provided signed coefficients ranked by magnitude, and Cubist models displayed their rule sets. Predictive adequacy and potential biases were assessed using observed-versus-predicted plots, residuals-versus-fitted diagnostics, and normal quantile-quantile (QQ) plots, with automated text annotations highlighting trends, heteroscedasticity, and tail departures. Together, these views provide complementary global and local insights while preserving the original training recipe and avoiding information leakage.

## Compatibility of Solvents and Polymers

The compatibility of solvents and polymers was studied to ensure that both the trained models and any simulated formulations remain chemically plausible and experimentally actionable, rather than artifacts of data-driven optimization. By explicitly encoding feasibility, predictions are aligned with laboratory practice, and reported results are made more trustworthy.

Chemical feasibility was enforced using two curated resources and integrated into both data preparation and simulation. First, a solvent–solvent incompatibility list was loaded, and solvent names were canonicalized before lookup. Pairwise incompatibility was checked symmetrically; any candidate mixture containing an incompatible pair was either rejected or simplified to a chemically plausible subset. Second, a polymer−solvent solubility table was loaded that accepts either a status or rating field and an optional maximum allowable percentage (max_pct). Ratings were standardized to OK, COND (conditionally allowed), or NO (not soluble), and solvent names were canonicalized in the same way as in the main data set to ensure consistent matching.

During prediction and Inverse Monte Carlo (IMC) workflows, a row-level solubility flag was computed by querying the polymer−solvent table for each of up to three solvents present. The aggregate rule assigned NO if any solvent was rated NO, COND if none was NO but at least one was COND, and OK otherwise. For mixtures generated in optimization mode, proposed solvent sets and their proportions were sampled within observed numeric ranges and then filtered by the compatibility rules. In particular, for polymer−solvent pairs labeled COND without a specified max_pct, three user-selectable strictness modes were implemented: Strict (conditional solvents effectively disallowed), Balanced (default heuristic cap of 20% of the solvent mixture), and Lax (default heuristic cap of 30%). These values are intended solely as conservative search parameters for exploring conditional solvent systems when solvent-specific limits are unavailable and should not be interpreted as universal thermodynamic solubility thresholds. Whenever solvent-specific maximum allowable proportions (max_pct) are available, those values take precedence. NO pairs were disallowed unless a small tolerance (no_allow_pct) had been explicitly specified. The acceptance rate of proposals was recorded to quantify how strongly the chemical constraints restricted the feasible design space. This mechanism ensured that all reported formulations, rankings, and simulated successes were chemically credible with respect to both solvent−solvent miscibility and polymer−solvent solubility.

Polymer−solvent solubility and solvent−solvent compatibility are inherently context-dependent (e.g., concentration, temperature, cosolvent effects). Consequently, SpinCastML implements conservative feasibility rules designed to prioritize experimental practicality over aggressive extrapolation. Compatibility ratings are incorporated as constraints during the inverse-design search, while user-selectable strictness modes provide a transparent and flexible means of exploring conditionally compatible systems at different levels of risk tolerance. To promote transparency, reproducibility, and future community refinement, the underlying compatibility tables are distributed with the software. These strictness modes are intended as configurable feasibility-screening tools that guide exploration of chemically plausible formulations and should not be interpreted as predictive models of thermodynamic stability, phase behavior, or electrospinnability.

## ■ INVERSE MONTE CARLO SIMULATIONS

Inverse Monte Carlo (IMC) was used to invert a desired fiber diameter into plausible electrospinning settings under chemical and operational constraints. Two modes were implemented: Experimental mode and Optimization mode. In Experimental mode, configurations $x^{(i)}$ were sampled directly from the empirical distribution of the data set restricted to the user-selected polymer (eq 9); if fewer than five observations were available, the full data set was used to preserve variability.

$$x^{(i)} \sim D_{emp}(Polymer) \tag{9}$$

With $D_{emp}$ the empirical joint distribution of all experimental parameters. Because Experimental mode samples only real historical experiments, all configurations are inherently chemically feasible, and no acceptance filtering is applied.

In Optimization mode, the algorithm generates synthetic electrospinning conditions within observed experimental ranges. For each numeric variable $x_j$ (e.g., voltage, concentration, flow rate), values are sampled uniformly between the minimum and maximum values observed for the selected polymer (eq 10):

$$x_j^{(i)} \sim U(x_{j,min}, x_{j,max}) \tag{10}$$

Categorical variables (e.g., collector type) are sampled from their observed frequencies. Solvent systems are generated by choosing one to three solvents drawn from the solvent pool for the selected polymer, and their proportions were sampled using a Dirichlet distribution, ensuring ratios sum to 100%.

As mentioned in previous subsection, each candidate solvent system was screened using (i) a symmetric solvent−solvent incompatibility matrix and (ii) a polymer−solvent solubility table containing categorical ratings (OK, COND, NO) and, when available, solvent-specific maximum allowable proportions (max_pct). For a given polymer−solvent pair (p, s) and fraction r, the solubility feasibility indicator was calculated as indicated in eq 11.

$$I_{ps}(p, s, r) = \begin{cases} I(r \le \text{no_allow_pct}), \text{ if } rating = NO, \\ I(r \le \text{max_pct}), \text{ if } rating \\ \quad = COND \text{ and } \max_pct \text{ available}, \\ I(r \le \text{thr_strict}), \text{ if } rating \\ \quad = COND \text{ and } \max_pct \text{ unavailable}, \\ 1, \text{ if } rating = OK \end{cases} \tag{11}$$

where the strictness thresholds were indicated in eq 12.

$$thr_strict = \begin{cases} 0\%, \text{ if } strict\ mode, \\ 20\%, \text{ if } balance\ mode, \\ 30\%, \text{ if } lax\ mode \end{cases} \tag{12}$$

Solvent−solvent miscibility was encoded similarly through a binary indicator $I_{ss}(x)$, equal to 1 only if all solvent pairs in the configuration were mutually compatible.

A simulated configuration is accepted if and only if all chemical constraints were satisfied the eq 13.

$$A^{(i)} = \mathbb{1}\left( I_{ss}(x^{(i)}) = 1 \wedge \prod_{j=1}^{k} I_{ps}(p, s_j, r_j^{(i)}) = 1 \right) \tag{13}$$

The acceptance rate, reported only in Optimization mode, quantified the stringency of the chemical constraints (eq 14).

$$\hat{P}_{acc} = \frac{1}{N} \sum_{i=1}^{N} A^{(i)} \tag{14}$$

For each accepted configuration, the training-time preprocessing pipeline was applied to guarantee identical scaling, encoding, and transformations. The fitted model produced a predicted fiber diameter given by eq 15.

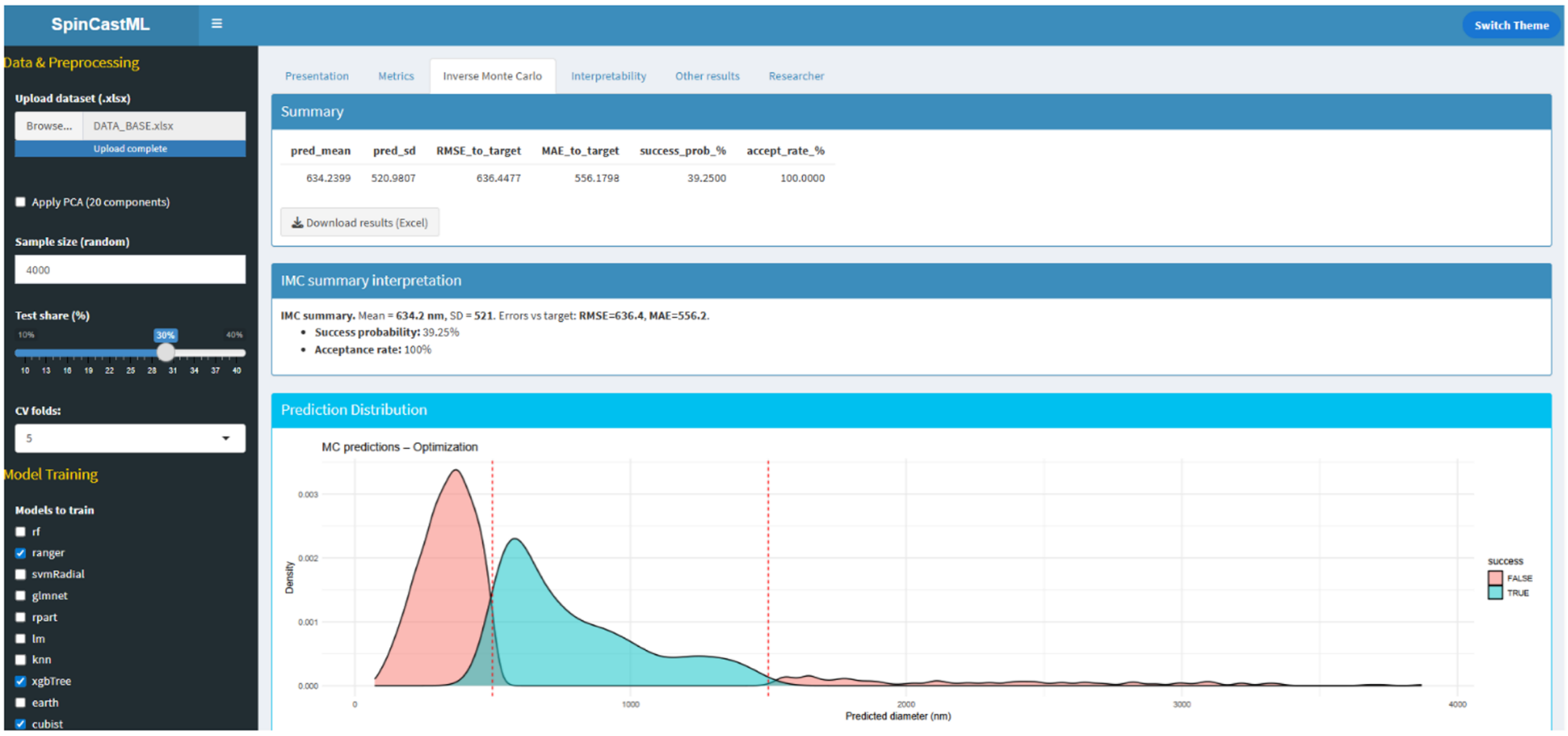

**Figure 11.** SpinCastML interface.

$$\hat{y}^{(i)} = \hat{f}\,(x^{(i)}) \tag{15}$$

Success was defined as achieving the target diameter $y^*$ within a user-defined tolerance $\varepsilon$ (eq 16):

$$|\hat{y}^{(i)} - y^*| \leq \varepsilon \tag{16}$$

With $\hat{y}^{(i)}$ the predicted diameter, $y^*$ the target diameter and $\varepsilon$ the tolerance.

Across a user-specified number of simulations, summary statistics were reported: the mean and standard deviation of predictions (pred_mean, pred_sd), error to target (RMSE_to_target, MAE_to_target), the estimated success probability (proportion within tolerance), and, for the Optimization mode, the proposal acceptance rate. Where the estimated success probability of meeting the tolerance band is determined by eq 17.

$$\hat{P}_{succ} = \frac{1}{N}\sum_{i=1}^{N} \mathbb{I}(|\hat{y}^{(i)} - y^*| \leq \varepsilon) \tag{17}$$

Each simulated row was additionally labeled with a solubility flag (OK/COND/NO) derived from the polymer−solvent table, enabling downstream ranking and filtering of candidate formulations. Overall, the IMC provided a principled, data-informed exploration of the formulation space while enforcing polymer−solvent solubility, solvent−solvent miscibility, and operational feasibility.

## SpinCastML Interface

Shiny, an R-based framework for interactive applications, was used to deliver SpinCastML, a desktop tool packaged as a self-contained Windows executable (.exe) that bundles the R runtime, required packages, data set and chemical interactions, and application code. The executable launches a local Shiny session, enabling fully offline operation while preserving reproducibility and auditability through version-locking of R/dependencies and archiving of the source. For a large codebase, this approach minimizes dependency drift, reduces user setup burden, improves performance predictability on local hardware, and strengthens data privacy by keeping all processing on the end user's machine.

SpinCastML is provided as a Shiny dashboard with a left-hand sidebar for inputs and a tabbed workspace for outputs. In Data & Preprocessing, an Excel data set (.xlsx) can be uploaded (large files are supported up to 200 MB), with immediate schema harmonization and solvent canonicalization performed in the background. Once a data set is loaded, the Polymer selector is populated dynamically from the data, and a small status line displays counts of solubility table entries by rating (OK/COND/NO) to confirm that chemical constraints have been ingested. Users can optionally apply principal component analysis (PCA; up to 20 components), constrain training time for large data sets by specifying a random sample size, a hybrid Sobol−D-optimal sampling, or a polymer-balanced Sobol and D-optimal sampling scheme, adjust the hold-out test proportion between 10% and 40% (default 30%), and select the number of cross-validation folds (3, 5, or 10). Under Model Training, candidate algorithms are chosen from a checklist; a single Train models action starts k-fold CV with a 70/30 split, showing a progress message while parallel training proceeds. The best model can be saved as a self-contained.RDS bundle (including the fitted model and the preprocessing recipe) and later reloaded to reproduce predictions without retraining. The Inverse Monte Carlo block exposes scientific controls: mode selection (Experimental vs Optimization), solubility strictness (Strict/Balanced/Lax), an optional tolerance for otherwise forbidden NO pairs (no_allow_pct), the target fiber diameter and tolerance window for fiber diameter, and the number of simulations. A compact theme switcher is provided both as a drop-down (Blue, Green, Dark) and as a floating Switch Theme button that applies background color changes instantly to favor inclusivity.

The main workspace is divided into tabs that mirror the analysis flow. The "Presentation" tab introduces the problem, renders the mathematical formulation of IMC, summarizes variables and metrics, and displays an illustrative electrospinning image; an IMC workflow diagram clarifies the pipeline from data preparation to outputs.

The "Metrics" tab reports the selected "best" model (by minimum RMSE on the test set), a full table of CV and test metrics (RMSE, MAE, MAPE, $R^2$) in an interactive DataTable, and a narrative explanation autogenerated from the results. Three diagnostic plots (Observed vs Predicted, Residuals vs Predicted, and a Residuals QQ-plot) are rendered with concise textual interpretations that flag bias, heteroscedasticity, or tail departures when detected.

The "Inverse Monte Carlo" tab surfaces simulation results. A summary table displays the mean and standard deviation of predicted diameters, error to target (RMSE and MAE), the estimated success probability (proportion within the target diameter of the fibers ± tolerance), and, in Optimization mode, the proposal acceptance rate after chemical vetoes. An accompanying density plot visualizes the prediction distribution with dashed lines marking the tolerance band. A top-20 table lists the closest chemically viable combinations to the target, retaining solvent identities and proportions, operational settings, predicted diameter, error metrics, and any available source links (DOI/paper URLs). Tables support horizontal scrolling and export of the full simulation set to Excel via a Download results (Excel) button.

Model explanation is gathered in the "Interpretability" tab. Global variable importance is plotted, a two-variable heat map explores interactions over observed ranges while conditioning other inputs at median/mode values, and an interactive 3-D surface provides an overview of the response landscape.

Additional artifacts are grouped in "Other results" tab: regression coefficients for linear and elastic-net models, a shallow surrogate decision tree that approximates model logic, human-readable Cubist rules when applicable, a compact Diagnostic table summarizing data/model dimensions and target variability, and a downloadable Full report (PDF) for archiving.

A Researcher tab introduce the author of SpinCastML and presents a reference diagram of the followed method.

All plots respect the preprocessing recipe attached to the active model to prevent leakage, and all interactive tables are delivered via DataTables (DT) for sorting, paging, and export. The design therefore supports end-to-end operation (data loading, model selection, chemically constrained simulation, interpretability, and reporting) within a single, coherent environment.

SpinCastML interface can be found in Figure 11.

### Executable Development

A self-contained desktop executable of SpinCastML was engineered to run without a preinstalled R environment by bundling a fixed R-portable runtime, the compiled package library, and all application assets into a relocatable directory. The codebase was organized around a single entry script (SpinCastML.R) with Shiny UI/server definitions, supported by static assets, template for data resources (e.g., base data set and solubility tables), and report.Rmd for automated reporting; version-locked R packages were preinstalled into the portable library to ensure offline operation and reproducibility. Execution was orchestrated by a lightweight launcher that starts Shiny on a loopback end point with an automatically selected free port, opens the system browser, and enforces a robust working directory so relative paths to assets resolve consistently. Distribution was kept simple. The app was packaged as a double-clickable launcher (run_app.bat). In all cases, the trained model and its preprocessing pipeline were saved together in a single.RDS file (R's native single-object format). The executable loads this bundle to analyze new data sets locally, no retraining required.

### Prospective Case Study

After model training and benchmarking, the selected Cubist model was used within the SpinCastML Inverse Monte Carlo module to perform a practical inverse-design case study using poly(vinyl alcohol) (PVA). The polymer was fixed as PVA, and the target fiber diameter was set to 250 nm with a tolerance of ±50 nm, corresponding to an acceptable design window of 200−300 nm.

The Inverse Monte Carlo search generated 20 candidate formulation−process families, ranked according to predicted proximity to the target diameter and probability of satisfying the tolerance window. These families were then screened to select one condition for experimental validation. The final selection was based on chemical feasibility, predicted diameter within the target range, compatibility with the available electrospinning equipment, and practical implementability of the suggested parameters.

Numerical outputs such as polymer concentration or voltage were rounded to experimentally realistic values before laboratory testing. When several candidate families remained feasible after rounding, the condition with the highest practical PVA concentration was prioritized to improve productivity, provided that the predicted fiber diameter remained within the target window.

Following the selected IMC-derived condition, PVA (#P8136, Sigma-Aldrich, UK) was dissolved in distilled water ($dH_2O$) at 12% w/w by heating to 80 °C under continuous stirring until a homogeneous solution was obtained. For electrospinning, 5 mL of the prepared solution was loaded into a 10 mL syringe and processed using an Inovenso NE300 electrospinning apparatus. The solution was delivered through a 22 G needle at 0.3 mL $h^{-1}$. A voltage of 25 kV was applied between the needle tip and the collector, and fibers were deposited onto aluminum foil attached to a flat collector positioned 10 cm from the needle tip. Electrospinning was carried out at room temperature (20 °C).

Fiber diameter (Ø) was quantified in AxioVision SE64 Rel. 4.9.1 (Carl Zeiss SMT Ltd., Cambridge, United Kingdom). To guarantee reproducibility, three replicates were performed on different days and two samples per replicate were evaluated. For each sample, 20 fibers were measured from representative high-magnification fields. The resulting fiber-diameter distribution was then compared with the SpinCastML target range of 250 ± 50 nm.

## ■ ASSOCIATED CONTENT

### Data Availability Statement

The online version contains available supplementary material to support the results and discussion section. The software developed in this study is released under the GNU GPL 3.0 license and is openly available as a citable archive at https://doi.org/10.5281/zenodo.18557989, including code, input files, instructions and license.

### Supporting Information

The Supporting Information is available free of charge at https://pubs.acs.org/doi/10.1021/acsomega.6c07333.

(PDF)

## ■ AUTHOR INFORMATION

### Corresponding Author

**Elisa Roldán** − *Department of Engineering, Faculty of Science & Engineering, Manchester Metropolitan University, Manchester M1 5GD, U.K.*; orcid.org/0000-0002-7793-7542; Email: Elisa.Roldan-Ciudad@mmu.ac.uk

### Author

**Tasneem Sabir** − *Manchester Fashion Institute, Faculty of Arts & Humanities, Manchester Metropolitan University, Manchester M15 6BG, U.K.*

Complete contact information is available at: https://pubs.acs.org/10.1021/acsomega.6c07333

### Author Contributions

Conception of the article, literature review, data curation, coding, data analysis, statistical and machine learning studies, software development, writing, and review of the article: E.R. Review of the article: T.S.

### Funding

This research received no external funding.

### Notes

The authors declare no competing financial interest.

## ■ ACKNOWLEDGMENTS

A preprint of this manuscript is available on arXiv at: 10.48550/arXiv.2602.09120. We confirm that Figure [1] and all graphical elements included in the manuscript are free domain/were created by the authors and do not contain any third-party copyrighted material.

## ■ REFERENCES